\pdfoutput=1
\documentclass[11pt]{article}

\usepackage{ACL2023}

\usepackage{times}
\usepackage{latexsym}

\usepackage[T1]{fontenc}
\usepackage[utf8]{inputenc}

\usepackage{microtype}

\usepackage{inconsolata}

\usepackage{graphicx}
\usepackage{multirow}
\usepackage{tablefootnote}
\usepackage{footnote}
\usepackage{footmisc}
\usepackage{subcaption}
\usepackage{url}
\usepackage{booktabs}
\usepackage{linguex}
\usepackage{amsmath}
\usepackage{amssymb}
\usepackage{fontawesome}
\usepackage{tikz-dependency}
\usepackage{adjustbox}
\usepackage{tabularx}

\newcommand\genre[1]{{\texttt{#1}}}

\title{GENTLE: A Genre-Diverse Multilayer Challenge Set \\ for English NLP and Linguistic Evaluation
}

\author{Tatsuya Aoyama, Shabnam Behzad, Luke Gessler, Lauren Levine, Jessica Lin,\\{\bf Yang Janet Liu, Siyao Peng, Yilun Zhu, Amir Zeldes} \\
        Corpling Lab \\ Georgetown University \\
        \{ta571, sb1796, lg876, lel76, yl1290, yl879, sp1184, yz565, az364\}@georgetown.edu}
\begin{document}
\maketitle
\begin{abstract}
We present GENTLE, a new mixed-genre English challenge corpus totaling 17K tokens and consisting of 8 unusual text types for out-of-domain evaluation: dictionary entries, esports commentaries, legal documents, medical notes, poetry, mathematical proofs, syllabuses, and threat letters. 
GENTLE is manually annotated for a variety of popular NLP tasks, including syntactic dependency parsing, entity recognition, coreference resolution, and discourse parsing.
We evaluate state-of-the-art NLP systems on GENTLE and find severe degradation for at least some genres in their performance on all tasks, which indicates GENTLE's utility as an evaluation dataset for NLP systems.

\end{abstract}

\section{Introduction}

In the past several years, there have been great advances in NLP system performance on various tasks. However, many of these tasks are still evaluated on in-domain data, i.e.~held-out data taken from the same domain as the system's training data. 
While this methodology is sound, it often overstates systems' ability to perform in real-world settings, where out-of-domain (OOD) data can lead to significant degradation \cite{plank2016non,joshi-etal-2018-extending}, even when target data comes from a similar domain \cite{nayak-etal-2020-domain}. 
For this reason, it is essential to have evaluation datasets with diverse text types, which can give a more accurate picture of systems' capabilities on OOD data, especially for domains that are distant from commonly studied domains or underrepresented in existing training datasets.

In this paper, we present GENTLE (\textbf{GEN}re \textbf{T}ests for \textbf{L}inguistic \textbf{E}valuation), a small but ``extreme'' open-access dataset that can be used for OOD evaluation of popular NLP tasks in English, as well as for linguistic analysis of less studied genres. 
The NLP tasks considered here include morphosyntactic tagging and dependency parsing according to Universal Dependencies (UD, \citealt{MarneffeEtAl2021}), nested named and non-named entity recognition (NNER), coreference resolution, entity linking (Wikification), and hierarchical discourse parsing in the framework of Rhetorical Structure Theory (RST, \citealt{MannThompson1988}).\footnote{The corpus is also openly released as part of the Universal Dependencies 2.12 version available at \url{https://github.com/UniversalDependencies/UD\_English-GUM}.} 
Our data comes from eight genres explicitly selected to represent unusual and diverse data types not currently included in the English Universal Dependencies corpora: dictionary entries, transcripts of live esports commentary, legal documents, medical notes, poetry, mathematical proofs, course syllabuses, and threat letters. 

GENTLE enables us to answer various questions, including how well state-of-the-art (SOTA) models can parse OOD data and whether or not OOD genres are equally difficult for all NLP tasks. 
Apart from NLP performance, we can also see whether the annotation tasks in our challenge genres are difficult for humans and how the difficulties that arise in individual genres in GENTLE differ from those in existing datasets.

The rest of this paper is structured as follows: Section \ref{sec:related-work} presents some related work on OOD testing. Section \ref{sec:corpus-description} presents an overview of the corpus, while Section \ref{sec:genre_variation} compares the genres in the corpus in detail. Section \ref{sec:eval} evaluates human agreement and NLP system performance on our data for each task, compared to more standard UD English data. Section \ref{sec:conclusion} offers our conclusions. Our corpus is available at \url{https://github.com/gucorpling/gentle}.

\section{Related Work}\label{sec:related-work}

Previous work has focused on the importance of genre diversity and OOD evaluation for many of the NLP tasks included in GENTLE, 
supporting the general conclusion that NLP system performance tends to degrade on OOD data.

In coreference resolution, \citet{moosavi-strube-2017-lexical} and \citet{zhu-etal-2021-ontogum} point out that existing models mainly rely on lexical features (e.g.~word embeddings) and may face the problem of overfitting because of the large overlap of vocabulary between training and testing data. Apart from overfitting, low recall resulting from domain discrepancy is another major problem for named entity recognition (NER, \citealt{AUGENSTEIN201761}). 

Despite a recent surge in approaches for discourse-level tasks, there is still room for improvement in this area, especially for OOD data \citep{atwell-etal-2021-discourse}. 
\citet{liu-zeldes-2023-eacl} investigate the impact of genre diversity in training data composition for RST discourse parsing, the task of recursively identifying relations between propositions. They show that diverse data is essential for stable and generalizable models for this task. 

Similarly to the present work, \citet{kanerva-ginter-2022-domain} conduct an OOD evaluation of Finnish dependency parsing, including constructing a relatively ``extreme'' OOD treebank, including 5 distinct genres (web documents, clinical, online discussions, tweets, and poetry).\footnote{\url{https://github.com/UniversalDependencies/UD_Finnish-OOD/}} Their experiments indicate that syntactic parsing performance degrades severely on OOD data, particularly on the LAS (labeled attachment score) metric. 

Data diversity is thus crucial for a range of NLP tasks, but the lack of diverse data available hampers training and evaluation. Previous corpus construction efforts cover a wide range of English genres, for example, 5 genres in the English Web Treebank (EWT, \citealt{silveira-etal-2014-gold}) for syntactic annotations, and 6 in OntoNotes \cite{WeischedelPradhanRamshawEtAl2012} for NER and coreference as well. However, both datasets lack nested, non-named entities, entity linking (Wikification), and discourse parsing. 
 
More recently, the UD English GUM corpus (Georgetown University Multilayer corpus, \citealt{Zeldes2017}), with data from 12 genres (academic articles, biographies, conversation transcripts, works of fiction, Reddit posts, how-to-guides, interviews, news articles, political speeches, textbook excerpts, Wikivoyage travel guides, and YouTube vlog transcripts), covers all of the annotations examined in this paper, and raises the expectation of being a possibly good training set for OOD targets, due to its  diverse content. 
Experiments in this paper will therefore use our newly annotated OOD GENTLE corpus to evaluate SOTA models trained on the already diverse GUM corpus and compare their performance on both datasets.

\section{GENTLE}\label{sec:corpus-description}

The GENTLE corpus is constructed as an OOD evaluation dataset, modeled on the test set for the English GUM corpus.
Table \ref{tab:partitions} gives an overview of partitions in GUM (v9.0) compared to GENTLE. 

\begin{table}[h]
\centering 
\small
\begin{tabular}{@{}lcll@{}}
\toprule
\textbf{dataset}      & \textbf{genres} & \textbf{docs} & \textbf{tokens}   \\ \midrule
GUM$_\textrm{train}$ & 12 & 165 & 160,700  \\
GUM$_\textrm{dev}$ & 12 & 24 & 21,409  \\
GUM$_\textrm{test}$ & 12 & 24 & 21,770  \\
GENTLE & 8 & 26 & 17,797  \\
\bottomrule
\end{tabular}%
\caption{GUM Partitions vs.~GENTLE. }
\label{tab:partitions}
\vspace{-10pt}
\end{table}

GENTLE forms an extension to the GUM test set with 8 more genres, for a total of 20 diverse text types to test on. Although the amount of data in GENTLE is small, the data follows GUM's scheme and is richly annotated on many layers, containing over 250K key-value annotations connected by complex annotation graphs. For treebanking, the annotations include gold-standard layers for Universal Dependencies morphosyntax, such as XPOS (Penn Treebank) tags, lemmas, and basic dependencies.
In addition, automatically-derived morphological features, enhanced dependencies and UPOS tags are obtained using the DepEdit library \cite{peng-zeldes-2018-roads} with the same scripts that produce these layers for the GUM corpus. 

For NNER and coreference resolution, the data includes nested, named and non-named entity annotations.
These employ the same scheme used in GUM, with 10 entity types, 6-way information status annotations, coreference and bridging links (9 edge types from GUM, including split antecedents, discourse deixis, etc., see \url{https://gucorpling.org/gum/}). 
GUM-style entity linking (wikification, \citealt{lin-zeldes-2021-wikigum}) is also provided, with an automatically produced alternate version of the entity/coreference annotations matching the OntoNotes scheme (\citealt{WeischedelPradhanRamshawEtAl2012}; see \citealt{zhu-etal-2021-ontogum} for details). 
The data also includes complete hierarchical discourse trees in Rhetorical Structure Theory (RST, \citealt{MannThompson1988}), following the same scheme as GUM.

Annotation was conducted by the authors of this paper during several hackathon-style annotation sessions. Although varying in expertise on each task, every annotator had previous experience annotating every layer of annotation described above. For annotation tools, morphosyntactic layers (XPOS tags, lemmas, and basic dependencies), entity layers (entity and coreference), and discourse layers (EDU segmentation and discourse relation) were annotated on Arborator \cite{gerdes-2013-collaborative} and Midas Loop \cite{gessler-etal-2022-midas}, GitDox \cite{zhang2017gitdox}, and rstWeb \cite{zeldes-2016-rstweb}, respectively. We also double annotated a portion of the corpus to measure human agreement, which will be further described in \S \ref{sec:eval}.

In choosing data, we attempted to select challenging types of spoken and written open-access materials that are maximally different from those already found in GUM (cf.~\S \ref{sec:related-work}). Texts were selected for each genre from a single source, making sure that (1) the total number of tokens falls between 2k and 2.5k tokens per genre; (2) at least 2 texts are selected to better represent the genre (as mean can only be calculated with 2 or more documents per genre). While the texts were selected randomly for most genres, the texts for some genres were manually selected. For instance,  since poetry texts can be extremely short, the documents for this genre were chosen to be varied in length, as to limit the number of documents needed to reach the target token range.  Table \ref{tab:overview-gentle-stats} gives the genre composition and sources for each data type in GENTLE.\footnote{Please refer to Appendix \ref{sec:genres} for detailed information on the contents of each genre.}

\begin{table}[ht]
%\resizebox{\columnwidth}{!}{%
\centering \small
\begin{tabular}{@{}lcll@{}}
\toprule
\textbf{genre}      & \multicolumn{1}{l}{\textbf{docs}} & \textbf{tokens} & \textbf{source}   \\ \midrule
\genre{dictionary} & 3                                 & 2,423          & Wiktionary        \\
\genre{esports}    & 2                                 & 2,149                 & YouTube           \\
\genre{legal}      & 2                                 & 2,288                & Wikisource / CUAD\tablefootnote{The \textbf{C}ontract \textbf{U}nderstanding \textbf{A}tticus \textbf{D}ataset (CUAD) v1 from the The Atticus Project \cite{hendrycks2021cuad}: \url{https://www.atticusprojectai.org/}} \\
\genre{medical}    & 4                                 & 2,164                 & MTSamples           \\
\genre{poetry}     & 5                                 & 2,090                 & Wikisource        \\
\genre{proof}      & 3                                 & 2,106                 & Proofwiki         \\
\genre{syllabus}   & 2                                 & 2,431                & GitHub           \\
\genre{threat}     & 5                                 & 2,146               & casetext          \\ \midrule
\textbf{total}      & 26  &17,797  &  \\ \bottomrule
\end{tabular}%
%$}
\caption{Corpus Contents of GENTLE.}
\label{tab:overview-gentle-stats}
\vspace{-8pt}
\end{table}

The chosen data is broad not only in domain, including medical, legal, and other technical areas, but also in medium (online linked resources such as Wiktionary data, spontaneous spoken esports commentary, and threat letters) and communicative intent (e.g.~poetry, syllabuses, and mathematical proofs). These genres can also be challenging for both humans and NLP models, as they diverge in various ways from standard training data and materials that guidelines are based on for each task. 

Before approaching a technical evaluation of how well humans can annotate these materials (inter-annotator agreement) and how NLP models score on them for each task, in the next section, we explore how the materials differ from genres in GUM descriptively, in text content and annotations.

\section{Variation across Genres}\label{sec:genre_variation}

\subsection{Summary Statistics}

Because the materials in GUM and GENTLE cover a vast range of text types, a quantitative view of variation in the data can provide a useful starting point in understanding what makes each genre unique. 
Although we could also devote as much attention to GUM genres, for space reasons, we will focus here on how each GENTLE genre is distinct from GUM and other genres (for more on GUM genres, see \citealt{zeldes-simonson-2016-different}).

Table \ref{tab:genre-metrics} gives an overview of some commonly used descriptive metrics to compare GENTLE genres to the GUM corpus average, as well as the score for GUM's \genre{news} genre, which can be taken as a stand-in for the standard language typically found in reference corpora, e.g. the Wall Street Journal \cite{MarcusSantoriniMarcinkiewicz1993}.
The \textcolor{red}{lowest} and \textcolor{blue}{highest} numbers in each metric are colored in \textcolor{red}{red} and \textcolor{blue}{blue}.

\begin{table}[t]
\centering \small
\begin{tabular}{lrrrrrr}
\toprule

\textbf{genre}    & \textbf{slen} & \textbf{pass} & \textbf{n/v}  & \textbf{ttr}  & \textbf{oov}  &  \textbf{sglt}  \\

\midrule
GUM & 20.16 & .07 & 2.36 & .4 & -- & .29  \\
GUM$_{news}$ & 22.52 & .12 & 3.34 & .45 & -- & .28  \\
\midrule
dictionary & 10.98 & .1 & 3.65 & .39 & .11 & \textbf{\textcolor{blue}{.49}}  \\
esports & 21.07 & \textbf{\textcolor{red}{.01}} & 1.48 & .36 & .08 & .24  \\
legal & 21.58 & .04 & 3.33 & .36 & .17 & .33  \\
medical & 11.21 & .15 & 4.31 & .46 & .22 & .32  \\
poetry & 17.7 & \textbf{\textcolor{red}{.01}} & 1.59 & \textbf{\textcolor{blue}{.53}} & .11 & .25  \\
proof & 15.63 & \textbf{\textcolor{blue}{.18}} & 5.14 & \textbf{\textcolor{red}{.25}} & \textbf{\textcolor{blue}{.24}} & \textbf{\textcolor{red}{.13}}  \\
syllabus & \textbf{\textcolor{red}{7.65}} & .12 & \textbf{\textcolor{blue}{5.34}} & .43 & \textbf{\textcolor{blue}{.24}} & .38  \\
threat & \textbf{\textcolor{blue}{24.25}} & .02 & \textbf{\textcolor{red}{1.3}} & .49 & \textbf{\textcolor{red}{.05}} & .28 
\\ \bottomrule
\end{tabular}%
%$}
\caption{Average sentence length (\texttt{slen}), passive ratio (\texttt{pass}), noun/verb ratio (\texttt{n/v}), type-token ratio (\texttt{ttr}), out-of-vocabulary ratio (\texttt{oov}), and singleton  ratio (\texttt{sglt}).}
\label{tab:genre-metrics}
\vspace{-8pt}
\end{table}

\begin{table*}[t!b]
% \resizebox{\textwidth}{!}{%
\centering
\small
\begin{tabular}{l|lr|lr|lr|lr}
\toprule
 & \multicolumn{2}{c|}{UPOS} & \multicolumn{2}{c|}{dependency relations}  & \multicolumn{2}{c|}{entity types}  & \multicolumn{2}{c}{discourse relations}    \\
 \midrule

GUM & PROPN & $\downarrow$-25.13 & dep & $\downarrow$-20.93 & org. & $\downarrow$-19.50 & joint & $\downarrow$-8.89  \\
GUM$_{news}$ & PROPN & $\uparrow$41.41 & flat & $\uparrow$21.74 & org. & $\uparrow$30.60 & attrib. & $\uparrow$12.14  \\
\midrule
dictionary & X & $\uparrow$21.16 & punct & $\uparrow$20.37 & abstract & $\uparrow$17.05 & org. & $\uparrow$5.72  \\
esports & ADV & $\uparrow$5.91 & parataxis & $\uparrow$16.68 & event & $\uparrow$\textcolor{red}{9.94} & eval. & $\uparrow$6.82  \\
legal & X & $\uparrow$18.37 & dep & $\uparrow$17.20 & org. & $\uparrow$12.62 & context & $\downarrow$\textcolor{red}{-3.69}  \\
medical & NOUN & $\uparrow$11.59 & nummod & $\uparrow$7.88 & substance & $\uparrow$11.02 & joint & $\uparrow$12.86  \\
poetry & PROPN & $\downarrow$\textcolor{red}{-7.07} & compound & $\downarrow$-5.68 & animal & $\uparrow$17.84 & mode & $\uparrow$5.50  \\
proof & SYM & $\uparrow$\textcolor{blue}{56.27} & dep & $\uparrow$10.92 & abstract & $\uparrow$\textcolor{blue}{39.23} & explan. & $\uparrow$8.46  \\
syllabus & X & $\uparrow$54.48 & dep & $\uparrow$\textcolor{blue}{46.31} & abstract & $\uparrow$25.69 & joint & $\uparrow$\textcolor{blue}{18.23}  \\
threat & PRON & $\uparrow$13.17 & punct & $\downarrow$\textcolor{red}{-6.98} & person & $\uparrow$12.91 & explan. & $\uparrow$6.43  \\

\bottomrule
\end{tabular}%
% }
\caption{Strongest Standardized $\chi^2$ Residual Label in 4 Layers for each Genre.}
\label{tab:resid}
% \vspace{-8pt}
\end{table*}

Most genres in GENTLE have substantially shorter sentences (\textbf{slen}) than the GUM average, with \genre{syllabus} having the lowest mean of 7.65 tokens, largely due to frequent bulleted or numbered lists of course topics, which are noun phrase fragments (e.g.~\textit{Week 3 - JavaScript Fundamentals}). The only genre with substantially above average sentence length is \genre{threat}, in which long and sometimes rambling justifications or elaborate consequences are often added to main sentences.

 Passivization (\textbf{pass}) is rare overall, except for medical texts (double the GUM average) and math proofs (even more), in which volitional agents are often suppressed (in the former, someone was \textit{diagnosed} but we do not know by whom; in the latter, a variable \textit{can be assigned}, etc.).

Noun/verb ratio (\textbf{n/v}) and type-token ratio (\textbf{ttr}) reveal that \genre{syllabus} has a rich and mainly nominal vocabulary (lists of skills or topics, primarily nouns/compounds). 
Though rich in \textbf{ttr}, \genre{threat} is more verbal.
\genre{poetry} has the highest \textbf{ttr}, partly because some poetic constraints discourage repetition (e.g.~alliteration and rhyming, where duplication is avoided).
In contrast, \genre{proof} has the lowest \textbf{ttr} since some terms are used repeatedly (e.g., \textit{vertex} is repeated ten times in one proof about vertices). 

The out-of-vocabulary (\textbf{oov}) rate shows the percentage of tokens in each genre that is not attested in GUM, which can be expected to correlate with NLP tool degradation.
\genre{proof}, \genre{syllabus} and \genre{medical} have extremely high rates (nearly 25\% of tokens are never seen in GUM), while \genre{threat} and \genre{esports} have less alarming rates of 5--8\%.

Finally, the proportion of singleton mentions (\textbf{sglt}, entities referred to just once in a text) shows that \genre{proof} documents have repetitive vocabularies and repeatedly refer to the same entity. 
This is because once a member or a class of possible items has been introduced, its properties are discussed in detail (e.g., after defining \textit{Let DE be a rational straight line}, we may continue discussing the line DE). By contrast, \genre{dictionary} documents use many arbitrary entities in example sentences that are never mentioned again (in an example sentence for \textit{school}, we find \textit{Harvard University is a famous American post-secondary school}, but \textit{Harvard} is then never mentioned again). These genre disparities and unique environments can be expected to interfere with prior probabilities learned by NLP models, and, as we will see below, also with human annotation agreement.

\subsection{Label Distributions}\label{subsec:label-distribution}

To give a quick overview of which labels deviate from their expected frequency in each genre, Table \ref{tab:resid} gives standardized chi-square residuals in a contingency table of labels versus genres. A positive residual means that a label is used more frequently than expected based on its overall frequency, and a negative residual means the opposite -- that a label is used less frequently than expected. Here we give only the strongest deviation associated with each genre in each of four annotation layers (for the complete tables of residuals, see Appendix \ref{appendix:residuals}). 

The deviation with the absolute highest score in the parts-of-speech (UPOS) is the unsurprising frequency of the tag \texttt{SYM} in math proofs, used for many mathematical symbols. The second highest is the tag \texttt{X} in \genre{syllabus}, used to tag bullet point markers and also used frequently in \genre{legal} documents. Other tag deviations include the lack of proper nouns in \genre{poetry}, dense use of punctuation in dictionary entries, and the prevalence of common nouns in the medical data (a lack of pronouns mirrors this, see Table \ref{tab:resid-upos} in Appendix \ref{appendix:residuals}). 

Dependencies show some parallel phenomena (\texttt{punct} in \genre{dictionary}, \texttt{dep} in \genre{legal} and \genre{syllabus}, which is used to attach bullet points), but also reveal lack of punctuation in threat letters. The prevalence of \texttt{parataxis} in \genre{esports} to narrate chains of events as they unfold is also noteworthy, as in \ref{ex:parataxis}, and the use of numerical quantities in medical texts, often used for medication dosages as in \ref{ex:nummod}. The \genre{poetry} genre shows a negative deviation in avoiding nominal compounds, which are more typically a property of technical texts in English, e.g.~in nested noun-noun compounds found in \genre{medical} notes, as in \ref{ex:compound}. 

\ex. \textit{Jovi\'{c} scoring, van de Beek and Ibrahimovic coming on 3-1 ... }\label{ex:parataxis}

\ex. \textit{Prilosec 20 mg b.i.d.} \label{ex:nummod}

\ex. \textit{white blood cell count} \label{ex:compound}

Residuals of entity types also expose differences compared to GUM genres and \genre{news} in particular, which distinguishes itself by frequently mentioning \texttt{organization} entities. \genre{proof} is the most extreme in favoring the \texttt{abstract} type (in fact, over 96\% of mentions in \genre{proof} are \texttt{abstract}), while \genre{threat} focuses on people. \genre{medical} is unique with its preponderance of \texttt{substance} entities, primarily medications, while \genre{esports} disproportionately uses the \texttt{event} type. One result in the table is an artifact of one specific document, and the small corpus size: \texttt{animal} in \genre{poetry} is due entirely to the inclusion of Edgar Alan Poe's ``The Raven''.

Finally, discourse relations reveal the prevalence of coordinated lists annotated in the relation class \textsc{joint} in \genre{syllabus} (topics, assignments, weeks in the course, etc.) and \genre{medical} (symptoms, vital statistics, medications; all mainly the relation subtype \textsc{joint-list}); \genre{esports} unsurprisingly favors \textsc{evaluation} to convey positive or negative impressions of players, and \genre{poetry} is unique in favoring \textsc{mode} relations, primarily due to the relation subtype \textsc{mode-manner}, which is used in adverbial manner adjuncts or parataxis, as in \ref{ex:manner1}--\ref{ex:manner2}. \genre{legal} shows a negative tendency to avoid \textsc{context} relations, which include background and spatio-temporal contextual information, both of which are less needed in a highly specialized and professional text in which context is often a given and statements apply in general. 

\ex. \textit{I sat divining, [with my head at ease]}$_{\textsc{manner}}$\label{ex:manner1}

\ex. \textit{[We slowly drove]}$_{\textsc{manner}}$ \textit{He knew no haste} \label{ex:manner2}

\subsection{Proximity across Genres}

The metrics in \S\ref{subsec:label-distribution} reveal differences among GENTLE genres compared to GUM.
But needless to say, there are also many similarities between  the GENTLE and GUM genres. To describe proximity across genres, we utilize the features in Table \ref{tab:genre-metrics} and the full residual tables for the four annotation layers in Appendix \ref{appendix:residuals} to build a cluster dendrogram of GENTLE and GUM genres. 

Because labels occupy different numerical ranges and have diverse tag set sizes (only 10 entity types but 34 coarse dependency labels), we scale the data by transforming it into z-scores, 

and then reduce the dimensionality of each table of residuals to five columns using Principal Component Analysis (PCA). In other words, while the original table of entity residuals has one row per genre and ten columns for the entity types (Table \ref{tab:resid-ent} in the Appendix) and contains chi-square residuals, the transformed table is based on a z-scaled version of the same table, which is reduced to having only 5 total columns using PCA.
This affords each annotation layer as much space as the five features in Table \ref{tab:genre-metrics} (excluding OOV rate, which is inapplicable to GUM data), for a total of 25 features per genre (the five scaled metrics without OOV, and five features each for POS, dependencies, entities, and discourse relations).

\begin{figure*}[t!b]
    \centering
    \includegraphics[scale=0.8,width=0.9\textwidth,clip,trim = 0cm 2cm 0cm 0cm]{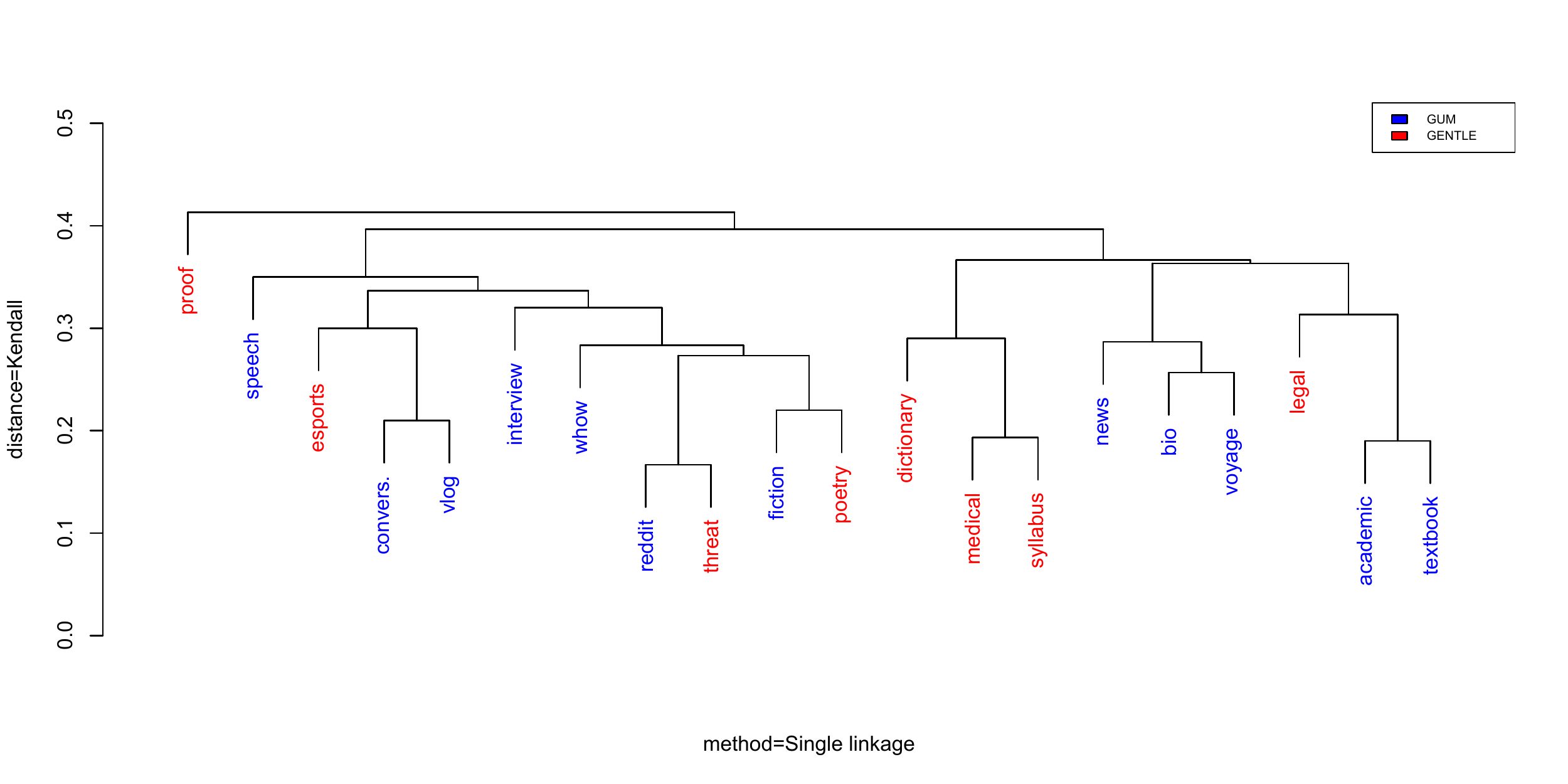}
    \caption{Cluster Dendrogram for GUM and GENTLE Genres.}
    \label{fig:clustering}
\end{figure*}

Because we are interested in concord/discord between genres across layers and do not necessarily care if z-scores are more or less extreme for a particular annotation layer, we use ordinal Kendall correlations between values of each dimension to compute the distance metric between genres, thereby avoiding single features with large values dominating the clustering. In the ordinal clustering, genres are closer if their ranks for multiple features are ordered more similarly---e.g., if they are ranked first and second in type-token ratio and singletons, then those two genres display positive concord along those features. We apply single linkage clustering to produce the dendrogram in Figure \ref{fig:clustering}.\footnote{An anonymous reviewer has inquired whether we attempted other clustering procedures: the answer is yes---the decision to use ordinal clustering resulted from the observation that single annotation layers had outsize influence for some genres, such as SYM tags in \genre{proof}; single linkage is both a default choice, and works well to cluster pairs of near genres as dendrogram leaves.}

As the figure shows, several of the GENTLE genres (in \textcolor{red}{red}) form outliers and cluster apart from genres in GUM (in \textcolor{blue}{blue}).
This suggests, on the one hand, they are substantially distinct and, therefore, valuable additions to already available genres in GUM. 
On the other hand, they may be challenging to handle for models trained on GUM. This is especially true for genres like \genre{proof} on the left side of the plot, which forms the most distinct outlier, in a top-level cluster of its own, and quite distant vertically from other genres. We can also see \genre{legal} quite distant from its nearest neighbors, GUM's \genre{academic} and \genre{textbook}, which are near each other. 
Three GENTLE genres, \genre{dictionary}, \genre{medical}, and \genre{syllabus} form a sub-cluster, with the latter two being relatively similar, possibly due to both genres being dominated by bulleted lists comprised of noun phrases, i.e.,~sentences fragments.

In the middle, \genre{poetry} is the closest to GUM's \genre{fiction}, perhaps partly due to long sentences, extensive vocabulary, and verb-dominated morphosyntax. 
\genre{threat} clusters with GUM's \genre{reddit} genre, perhaps because both are relatively argumentative genres, often written in first-person, and include many interjections and swearwords (see \citealt{behzad-zeldes-2020-cross} for similar and additional observations on Reddit data). \genre{esports} is somewhat far from its nearest neighbors, the informal spoken genres \genre{conversation} and \genre{vlog} which intuitively share features and cluster together; the latter also comes from the same source and modality as \genre{esports}, since both were collected from YouTube. GUM's more informative expository genres also cluster together plausibly, with biographies (\genre{bio}) and travel guides (\genre{voyage}) grouped together after the split with \genre{news}.

\begin{table*}[t]
    \centering
    \small
    \resizebox{\textwidth}{!}{%
\begin{tabular}{@{}llcccccccccc@{}}
\toprule
\textbf{Tasks} &
  \textbf{Metrics} &
  \multicolumn{1}{c}{\textsc{\textbf{micro}}} &
  \multicolumn{1}{c}{\textsc{\textbf{macro}}} &
  \multicolumn{1}{c}{\genre{\textbf{dictionary}}} &
  \multicolumn{1}{c}{\genre{\textbf{esports}}} &
  \multicolumn{1}{c}{\genre{\textbf{legal}}} &
  \multicolumn{1}{c}{\genre{\textbf{medical}}} &
  \multicolumn{1}{c}{\genre{\textbf{poetry}}} &
  \multicolumn{1}{c}{\genre{\textbf{proof}}} &
  \multicolumn{1}{c}{\genre{\textbf{syllabus}}} &
  \multicolumn{1}{c}{\genre{\textbf{threat}}}\\ \midrule
  \multicolumn{12}{c}{\textit{\textbf{Human Agreement on Snippets}}}  \\
\midrule
\textbf{POS Tagging    } &    Acc &  95.38 &  95.37 & 94.69 &  \textcolor{blue}{\textbf{98.25}} &  \textcolor{red}{\textbf{93.48}} & 94.81 & 97.85 &  95.67 & 93.86 &  94.37 \\
\textbf{(\textsc{xpos})} &        $\kappa$ &  94.98 &  94.78 & 95.38 &   \textcolor{blue}{\textbf{98.00}} & 93.46 & 94.49 & 97.29 &  94.38 &  \textcolor{red}{\textbf{92.08}} &  92.94 \\

\midrule
\textbf{Lemmatization  } &    Acc &  96.90 &  96.89 &  \textcolor{red}{\textbf{92.92}} &  \textcolor{blue}{\textbf{99.56}} & 95.22 & 96.54 & 97.42 &   98.70 & 95.61 &  99.13 \\
       &        $\kappa$ &  96.86 &  96.82 &  \textcolor{red}{\textbf{92.65}} &  \textcolor{blue}{\textbf{99.55}} & 95.12 & 96.47 & 97.36 &  98.66 & 95.56 &  99.11 \\
\midrule
\textbf{Dependency     } &    UAS &  88.79 &  88.77 &  \textcolor{red}{\textbf{77.88}} &  85.53 &  90.00 & 88.74 & 90.13 &  88.74 & 93.86 &  \textcolor{blue}{\textbf{95.24}} \\
\textbf{Parsing        } &    LAS &  84.66 &  84.63 &  \textcolor{red}{\textbf{73.01}} &  81.58 & 83.48 & 87.01 & 88.41 &  83.55 & 89.47 &  \textcolor{blue}{\textbf{90.48}} \\
\midrule
\textbf{Entity         } &      P &  89.47 &  89.25 & 93.24 &  92.54 & 91.94 & 79.71 &  \textcolor{red}{\textbf{78.08}} &  \textcolor{blue}{\textbf{96.19}} & 86.36 &  95.71 \\
\textbf{Recognition    } &      R &  85.27 &  84.84 & 81.18 &  92.54 & 79.17 &  \textcolor{red}{\textbf{77.46}} & 82.61 &  \textcolor{blue}{\textbf{97.12}} & 84.44 &  83.75 \\
\textbf{(untyped)      } &      F &  87.32 &  86.88 & 86.79 &  92.54 & 85.07 &  \textcolor{red}{\textbf{78.57}} & 80.28 &  \textcolor{blue}{\textbf{96.65}} & 85.39 &  89.33 \\
\midrule
\textbf{Entity         } &      P &  81.91 &  81.35 & 90.54 &   \textcolor{red}{\textbf{70.15}} & 90.32 & 73.91 & 76.71 &  \textcolor{blue}{\textbf{96.19}} & 73.86 &  78.57 \\
\textbf{Recognition    } &      R &  78.06 &  77.32 & 78.82 &  70.15 & 77.78 & 71.83 & 81.16 &  \textcolor{blue}{\textbf{97.12}} & 72.22 &   \textcolor{red}{\textbf{68.75}} \\
\textbf{(typed)        } &      F &  79.94 &  79.19 & 84.28 &   \textcolor{red}{\textbf{70.15}} & 83.58 & 72.86 & 78.87 &  \textcolor{blue}{\textbf{96.65}} & 73.03 &  73.33 \\
\midrule
       &    MUC &  70.46 &  66.01 & 47.05 &  \textcolor{blue}{\textbf{94.44}} & 72.22 & 60.86 & 62.06 &  70.58 &  \textcolor{red}{\textbf{38.09}} &  82.75 \\
\textbf{Coreference    } &  B$^3$ &  77.63 &  77.21 &  83.50 &  \textcolor{blue}{\textbf{90.29}} & 75.31 & 65.65 &  \textcolor{red}{\textbf{62.97}} &  84.74 & 76.38 &  78.87 \\
\textbf{Resolution     } &  CEAF$_{\phi4}$ &  72.25 &  70.55 & 84.43 &  \textcolor{blue}{\textbf{86.38}} &  69.30 & 63.55 &   \textcolor{red}{\textbf{48.10}} &   74.50 & 73.46 &   64.70 \\
       & Avg. F &  73.45 &  71.26 & 71.66 &  \textcolor{blue}{\textbf{90.37}} & 72.28 & 63.35 &  \textcolor{red}{\textbf{57.71}} &  76.61 & 62.64 &  75.44 \\
\midrule
\multicolumn{12}{c}{\textit{\textbf{NLP Performance on Snippets}}}  \\
\midrule
\textbf{XPOS} & Acc & 92.56 & 92.55 & 86.73 & 97.66 & 95.36 & 97.55 & \textbf{\textcolor{blue}{97.71}} & 
\textbf{\textcolor{red}{77.63}} & 93.27 & 94.52 \\
\midrule
\textbf{Lemmatization} & Acc & 96.32 & 96.33 & 97.64 & \textbf{\textcolor{blue}{99.56}} & 97.10 & 96.25 & \textbf{\textcolor{red}{92.56}} & 94.81 & 94.44 & 98.27 \\
\midrule
\textbf{Dependency} & UAS & 80.69 & 80.65 & 65.34 & 85.23 & 87.83 & 87.01 & \textbf{\textcolor{blue}{90.41}} & \textbf{\textcolor{red}{54.69}} & 85.09 & 89.61 \\
\textbf{Parsing} & LAS & 76.22 & 76.18 & 59.00 & 79.39 & 82.75 & 83.41 & \textbf{\textcolor{blue}{87.55}} & \textbf{\textcolor{red}{50.65}} & 81.14 & 85.57 \\
\midrule
\textbf{Entity} & P & 75.63 & 75.14 & 72.22 & 70.42 & \textcolor{red}{\textbf{66.89}} & 74.86 & 72.80 & \textcolor{blue}{\textbf{84.91}} & 78.42 & 80.60 \\
\textbf{Recognition} & R & 70.01 & 69.81 & \textcolor{red}{\textbf{60.61}} & 64.88 & 61.72 & 71.21 & 69.80 & 73.33 & 71.40 & \textcolor{blue}{\textbf{78.26}} \\
 \textbf{(typed)} & F & 72.71 & 72.34 & 65.91 & 67.53 & \textcolor{red}{\textbf{64.20}} & 72.98 & 71.27 & \textbf{\textcolor{blue}{82.67}} & 74.74 & 79.41 \\
\midrule

 & MUC & 65.66 & 54.86 & \textbf{\textcolor{red}{0.00}} & \textbf{\textcolor{blue}{83.72}} & 30.30 & 80.95 & 74.62 & 52.30 & 42.85 & 74.15\\
\textbf{Coreference} & B$^3$ & 41.25 & 36.72 & \textbf{\textcolor{red}{4.49}} & 54.27 & 22.78 & 38.73 & 56.33 & 26.45 & 29.47 & \textbf{\textcolor{blue}{61.23}}\\
\textbf{Resolution   } & CEAF$_{\phi4}$ & 17.72 & 18.31 & \textbf{\textcolor{red}{1.80}} & 22.00 & 20.95 & 6.82 & \textbf{\textcolor{blue}{36.13}} & 14.32 & 15.79 & 28.67\\
 & Avg. F & 41.54 & 36.63 & \textbf{\textcolor{red}{2.10}} & 53.33 & 24.68 & 42.17 & \textbf{\textcolor{blue}{55.69}} & 31.02 & 29.37 & 54.68\\
\bottomrule
\end{tabular}%
}

    \caption{Human Performance and Corresponding NLP Performance on GENTLE Snippets for 5 NLP Tasks. The highest scoring (`easiest') GENTLE genres are highlighted in \textbf{\textcolor{blue}{blue}}, and the lowest scoring are in \textbf{\textcolor{red}{red}}.}
    \label{tab:humanEval}
\end{table*}

\section{Evaluation}\label{sec:eval}

\begin{table*}[ht]
\centering
\small
\resizebox{\textwidth}{!}{%
\begin{tabular}{@{}llcccccccccccc@{}}
\toprule

\textbf{Tasks} &
  \textbf{Metrics} &
  \textbf{GUM$_\textrm{test}$} &
  \textbf{GUM$_\textrm{test-news}$} &
  \multicolumn{1}{l}{\textbf{\begin{tabular}[c]{@{}c@{}}GENTLE\\ (\textsc{micro})\end{tabular}}} &
  \textbf{\begin{tabular}[c]{@{}c@{}}GENTLE\\ (\textsc{macro})\end{tabular}} &
  \genre{\textbf{dictionary}} &
  \genre{\textbf{esports}} &
  \genre{\textbf{legal}} &
  \genre{\textbf{medical}} &
  \genre{\textbf{poetry}} &
  \genre{\textbf{proof}} &
  \genre{\textbf{syllabus}} &
  \genre{\textbf{threat}} \\ \midrule
  
\textbf{Tokenization} &
  F & \textbf{99.61} & 99.67 & 97.29 & 97.46 & 98.12 & 99.52 & 95.55 & 97.73 & 99.59 & 97.98 & \textcolor{red}{\textbf{91.46}} & \textcolor{blue}{\textbf{99.69}}
   \\ \midrule
   
\textbf{\begin{tabular}[c]{@{}l@{}}POS Tagging\\ (\textsc{xpos})\end{tabular}} &
  Acc &  \textbf{97.46} & \textbf{97.85} & 88.34 & 88.56 & 90.74 & \textcolor{blue}{\textbf{95.89}} & 89.71 & 92.93 & 91.51 & 78.76 & \textcolor{red}{\textbf{75.22}} & 93.74
   \\ \midrule
   
% \textbf{\begin{tabular}[c]{@{}l@{}}POS Tagging\\ (\textsc{upos})\end{tabular}} &
%   Acc & \textbf{97.49} & \textbf{97.85} & 86.51 & 89.21 & 90.56 & \textcolor{blue}{\textbf{95.61}} & 90.45 & 93.45 & 92.57 & 78.83 & \textcolor{red}{\textbf{78.25}} & 93.98
%    \\ \midrule

\textbf{Lemmatization} & Acc & \textbf{98.13} & \textbf{98.52} & 92.38 & 92.64 & 95.53 & \textcolor{blue}{\textbf{98.29}} & 91.72 & 93.01 & 95.51 & 91.06 & \textcolor{red}{\textbf{79.74}} & 96.23 \\ \midrule
   
\textbf{\begin{tabular}[c]{@{}l@{}}Dependency\\ Parsing\end{tabular}} &
\begin{tabular}[c]{@{}l@{}}UAS\\ LAS\end{tabular} &
  \begin{tabular}[c]{@{}c@{}}\textbf{89.49}\\ \textbf{87.21}\end{tabular} &
  \begin{tabular}[c]{@{}c@{}}\textbf{89.68}\\ \textbf{87.45}\end{tabular} &
  \begin{tabular}[c]{@{}c@{}}76.71\\ 72.38\end{tabular} &
  \begin{tabular}[c]{@{}c@{}}77.01\\ 72.65\end{tabular} &
  \begin{tabular}[c]{@{}c@{}}75.39\\ 70.78\end{tabular} &
  \begin{tabular}[c]{@{}c@{}}83.99\\ 78.84\end{tabular} &
  \begin{tabular}[c]{@{}c@{}}77.23\\ 73.95\end{tabular} &
  \begin{tabular}[c]{@{}c@{}}81.15\\ 77.64\end{tabular} &
  \begin{tabular}[c]{@{}c@{}}76.74\\ 71.70\end{tabular} &
  \begin{tabular}[c]{@{}c@{}}71.23\\ 65.58\end{tabular} &
  \begin{tabular}[c]{@{}c@{}}\textcolor{red}{\textbf{63.99}}\\ \textcolor{red}{\textbf{59.94}}\end{tabular} &
  \begin{tabular}[c]{@{}c@{}}\textcolor{blue}{\textbf{86.37}}\\ \textcolor{blue}{\textbf{82.77}}\end{tabular} 
  \\ \midrule

\textbf{\begin{tabular}[c]{@{}l@{}}Entity \\ Recognition\\(typed)\end{tabular}} &
  \begin{tabular}[c]{@{}l@{}}P\\ R\\ F\end{tabular} &
  \begin{tabular}[c]{@{}c@{}}\textbf{77.14}\\ \textbf{76.24}\\ \textbf{76.88}\end{tabular} &
  \begin{tabular}[c]{@{}c@{}}65.01\\ 72.69\\ 68.64\end{tabular} &
  \begin{tabular}[c]{@{}c@{}}75.63\\ 70.01\\ 72.71\end{tabular} &
  \begin{tabular}[c]{@{}c@{}}75.10\\ 69.77\\ 72.30\end{tabular} &
  \begin{tabular}[c]{@{}c@{}}72.22\\ \textcolor{red}{\textbf{60.61}}\\ 65.91\end{tabular} &
  \begin{tabular}[c]{@{}c@{}}70.42\\ 64.88\\ 67.53\end{tabular} &
  \begin{tabular}[c]{@{}c@{}}\textcolor{red}{\textbf{66.56}}\\ 61.41\\ \textcolor{red}{\textbf{63.88}}\end{tabular} &
  \begin{tabular}[c]{@{}c@{}}74.86\\ 71.21\\ 72.98\end{tabular} &
  \begin{tabular}[c]{@{}c@{}}72.80\\ 69.80\\ 71.27\end{tabular} &
  \begin{tabular}[c]{@{}c@{}}\textcolor{blue}{\textbf{84.91}}\\ \textcolor{blue}{\textbf{80.56}}\\ \textcolor{blue}{\textbf{82.67}}\end{tabular} &
  \begin{tabular}[c]{@{}c@{}}78.42\\ 71.40\\ 74.74\end{tabular} &
  \begin{tabular}[c]{@{}c@{}}80.60\\ 78.26\\ 79.41\end{tabular} \\ \midrule

\textbf{\begin{tabular}[c]{@{}l@{}}Coreference \\ Resolution\end{tabular}} &
  \begin{tabular}[c]{@{}l@{}}MUC\\ B$^3$\\ CEAF$_{\phi4}$ \\ Avg. F \end{tabular} &
  \begin{tabular}[c]{@{}c@{}}\textbf{76.38}\\ \textbf{64.71}\\ \textbf{57.15} \\ \textbf{66.08}\end{tabular} &
  \begin{tabular}[c]{@{}c@{}}59.67\\ 53.97\\ \textbf{53.06} \\ 55.57\end{tabular} &
  \begin{tabular}[c]{@{}c@{}}60.89\\ 33.37\\  9.75\\ 34.67\end{tabular} &
  \begin{tabular}[c]{@{}c@{}}55.98\\ 33.91\\  11.18\\ 33.69\end{tabular} &
  \begin{tabular}[c]{@{}c@{}}\textcolor{red}{\textbf{9.30}}\\ \textcolor{red}{\textbf{14.74}}\\ \textcolor{red}{\textbf{4.91}}\\ \textcolor{red}{\textbf{9.65}}\end{tabular} &
  \begin{tabular}[c]{@{}c@{}}67.84\\ 45.49\\ \textcolor{blue}{\textbf{17.48}}\\ 43.60\end{tabular} &
  \begin{tabular}[c]{@{}c@{}}59.14\\ 31.07\\ 9.22\\ 33.14\end{tabular} &
  \begin{tabular}[c]{@{}c@{}}70.13\\ 32.78\\ 7.06\\ 36.66\end{tabular} &
  \begin{tabular}[c]{@{}c@{}}70.92\\ 43.98\\ 15.10\\ 43.33\end{tabular} &
  \begin{tabular}[c]{@{}c@{}}48.95\\ 29.08\\ 13.48\\ 30.50\end{tabular} &
  \begin{tabular}[c]{@{}c@{}}41.09\\ 20.88\\ 7.78\\ 23.25\end{tabular} &
  \begin{tabular}[c]{@{}c@{}}\textcolor{blue}{\textbf{80.48}}\\ \textcolor{blue}{\textbf{53.29}}\\ 14.38\\ \textcolor{blue}{\textbf{49.38}}\end{tabular} \\ \midrule
   
\textbf{\begin{tabular}[c]{@{}l@{}}RST EDU\\ Segmentation\\ (Gold)\end{tabular}} &
 \begin{tabular}[c]{@{}c@{}}P\\ R\\ F\end{tabular} & 
  \begin{tabular}[c]{@{}c@{}}\textbf{96.43}\\ \textbf{95.85}\\ \textbf{96.14}\end{tabular} & 
 \begin{tabular}[c]{@{}c@{}}\textbf{95.68}\\ \textbf{97.17}\\ \textbf{96.42}\end{tabular} & 
  \begin{tabular}[c]{@{}c@{}}93.90\\ 93.17\\ 93.53\end{tabular} & 
 \begin{tabular}[c]{@{}c@{}}93.21\\ 92.07\\ 92.60\end{tabular} & 
 \begin{tabular}[c]{@{}c@{}}\textcolor{blue}{\textbf{97.58}}\\ 95.48\\ 96.52\end{tabular} & 
 \begin{tabular}[c]{@{}c@{}}95.71\\ \textcolor{red}{\textbf{87.01}}\\ 91.16\end{tabular} & 
 \begin{tabular}[c]{@{}c@{}}90.07\\ 96.11\\ 92.99\end{tabular} & 
 \begin{tabular}[c]{@{}c@{}}\textcolor{blue}{\textbf{97.58}}\\ 96.58\\ \textcolor{blue}{\textbf{97.07}}\end{tabular} & 
 \begin{tabular}[c]{@{}c@{}}91.30\\ 88.06\\ 89.66\end{tabular} & 
 \begin{tabular}[c]{@{}c@{}}\textcolor{red}{\textbf{88.81}}\\ 87.89\\ \textcolor{red}{\textbf{88.35}}\end{tabular} & 
 \begin{tabular}[c]{@{}c@{}}94.35\\ \textcolor{blue}{\textbf{98.04}}\\ 96.16\end{tabular} & 
 \begin{tabular}[c]{@{}c@{}}93.62\\ 92.69\\ 93.16\end{tabular} \\ \midrule

\textbf{\begin{tabular}[c]{@{}l@{}}RST EDU\\ Segmentation\\ (Trankit)\end{tabular}} &
  \begin{tabular}[c]{@{}l@{}}P\\ R\\ F\end{tabular} &
   \begin{tabular}[c]{@{}c@{}}\textbf{93.63}\\ \textbf{93.48}\\ \textbf{93.55}\end{tabular} & 
 \begin{tabular}[c]{@{}c@{}}\textbf{92.91}\\ \textbf{96.46}\\ \textbf{94.65}\end{tabular} & 
  \begin{tabular}[c]{@{}c@{}}89.90\\ 86.78\\ 88.31\end{tabular} & 
 \begin{tabular}[c]{@{}c@{}}90.17\\ 87.79\\ 88.89\end{tabular} & 
 \begin{tabular}[c]{@{}c@{}}95.48\\ 86.24\\ 90.62\end{tabular} & 
 \begin{tabular}[c]{@{}c@{}}94.37\\ 87.01\\ 90.54\end{tabular} & 
 \begin{tabular}[c]{@{}c@{}}85.29\\ 92.23\\ 88.62\end{tabular} & 
 \begin{tabular}[c]{@{}c@{}}\textcolor{blue}{\textbf{97.92}}\\ \textcolor{blue}{\textbf{96.58}}\\ \textcolor{blue}{\textbf{97.24}}\end{tabular} & 
 \begin{tabular}[c]{@{}c@{}}87.46\\ 85.48\\ 86.46\end{tabular} & 
 \begin{tabular}[c]{@{}c@{}}87.41\\ 88.93\\ 88.16\end{tabular} & 
 \begin{tabular}[c]{@{}c@{}}\textcolor{red}{\textbf{80.48}}\\ \textcolor{red}{\textbf{73.48}}\\ \textcolor{red}{\textbf{76.82}}\end{tabular} & 
 \begin{tabular}[c]{@{}c@{}}92.98\\ 92.36\\ 92.67\end{tabular} \\ \midrule

\textbf{\begin{tabular}[c]{@{}l@{}}RST \\ Parsing\end{tabular}} &
  \begin{tabular}[c]{@{}l@{}}S\\ N\\ R\end{tabular} &
  \begin{tabular}[c]{@{}c@{}}\textbf{70.07}\\ \textbf{56.90}\\ \textbf{49.57}\end{tabular} &
  \begin{tabular}[c]{@{}c@{}}\textbf{71.89}\\ \textbf{60.61}\\ \textbf{56.40}\end{tabular} &
  \begin{tabular}[c]{@{}c@{}}62.15\\ 47.63\\ 37.64\end{tabular} &
  \multicolumn{1}{c}{\begin{tabular}[c]{@{}c@{}}62.83\\ 48.05\\ 38.16\end{tabular}} &
  \begin{tabular}[c]{@{}c@{}}59.31\\ 47.47\\ 30.52\end{tabular} &
  \begin{tabular}[c]{@{}c@{}}\textcolor{red}{\textbf{55.77}}\\ \textbf{\textcolor{red}{40.41}}\\ \textbf{\textcolor{red}{29.30}}\end{tabular} &
  \begin{tabular}[c]{@{}c@{}}\textcolor{blue}{\textbf{72.72}}\\ \textcolor{blue}{\textbf{59.79}}\\ \textbf{\textcolor{blue}{51.48}}\end{tabular} &
  \begin{tabular}[c]{@{}c@{}}65.51\\ 50.35\\ 46.88\end{tabular} &
  \begin{tabular}[c]{@{}c@{}}59.78\\ 40.87\\ 30.93\end{tabular} &
  \begin{tabular}[c]{@{}c@{}}69.11\\ 55.25\\ 41.73\end{tabular} &
  \begin{tabular}[c]{@{}c@{}}57.13\\ 44.18\\ 40.17\end{tabular} &
  \begin{tabular}[c]{@{}c@{}}63.29\\ 46.06\\ 34.23\end{tabular} \\ \bottomrule

\end{tabular}%
}

\caption{End-to-End NLP Performance on All Tasks on Full Plain Texts (averaged over 3 runs). Top and bottom scoring GENTLE genres are marked in \textbf{\textcolor{blue}{blue}} and \textbf{\textcolor{red}{red}} (GUM scores are nearly always higher, in \textbf{bold}).}
\label{tab:EVAL4NLP}
\end{table*}
\vspace{-5pt}

To understand how challenging GENTLE data is for both NLP models and humans, we evaluate representative systems on each task using the entire corpus and conduct an inter-annotator agreement (IAA) experiment by double annotating 10\% of the data. 
Table \ref{tab:humanEval} reports Cohen's Kappa ($\kappa$) and task-specific scores where applicable, taking the gold standard release data as a reference, compared to a second human's annotation. 
The double annotations were done without additional validation checks; in other words, the final gold data, subjected to stringent validations by the official UD validator and validation scripts from the English GUM repository, can be expected to be more consistent and reliable. Double annotated data comes from document initial ``snippets'' in each genre since non-initial sections may be incoherent for layers such as coreference. Each snippet was around 200-250 tokens in length, amounting to 1,838 tokens in total ($\approx$10.34\% of the entire corpus).

However, it is also true that NLP accuracy in document-initial positions diverges from overall accuracy since documents are systematically non-homogeneous. For example, \genre{dictionary} entry beginnings are much harder to parse since they contain technical notation, foreign language etymologies, and more, while later sections typically include grammatically simple usage example sentences. 
Therefore, we report NLP accuracy on the double annotated snippets compared to human scores in Table \ref{tab:humanEval}, separately from the overall performances on the GENTLE corpus in Table \ref{tab:EVAL4NLP}. For each setting, we report scores by genre, for the entire corpus (micro-average), and the averaged per-genre score (macro-average). All NLP models were trained on the GUM v9 train partition and tested on the established GUM v9 test set and GENTLE. Additionally, we include genre-specific numbers for GUM's \genre{news} section, which can be taken to represent the most commonly used evaluation data type in most NLP tasks.

\paragraph{Tokenization, Tagging, Lemmatization, and Dependency Parsing} 
We use the widely employed Stanza package \cite{qi-etal-2020-stanza} to evaluate gold-tokenized texts in Table \ref{tab:humanEval} allowing comparisons with human agreements, as well as end-to-end from plain text in Table \ref{tab:EVAL4NLP} to also evaluate tokenization.
Tokenization degrades in the end-to-end scenario for all GENTLE genres except for \genre{threat}.
Tokenization is error-prone in \genre{syllabus} and \genre{legal} due to the abundance of bulleted and numbered nominal phrases and abbreviations. 
XPOS tagging degrades nearly 10 points on GENTLE and scores the lowest on \genre{proof} and \genre{syllabus} due to mathematical symbols (e.g.~$\leqq$, $\in$, $x$, $y$) and genre-specific terminologies (e.g.~TAs, TBD).
Micro-averaged lemmatization performance drops nearly 6 points to 92.38 and parsing by 15 points to a LAS of 72.38, again worst for \genre{proof} and \genre{syllabus}. 

\begin{figure}[htb]
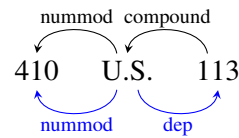

\centering
\begin{dependency}[arc edge, arc angle=80, text only label, label style={above}]
\begin{deptext}[column sep=.1cm]
410 \& \hspace{0.5em} U.S. \hspace{0.5em} \& 113 \\
%\lbrack sic\rbrack \&  \& \&  \\
\end{deptext}
\depedge{2}{1}{nummod}
\depedge{3}{2}{compound}
\depedge[edge below, edge style={blue}, label style={text=blue, below}]{2}{1}{nummod}
\depedge[edge below, edge style={blue}, label style={text=blue, below}]{2}{3}{dep}
\end{dependency}
\caption{Annotation Disagreement for \textit{410 U.S. 113}.}\label{fig:us113}
\vspace{-10pt}
\end{figure}

While these results may be somewhat shocking, human performance is also imperfect, with XPOS and lemmatization accuracy in the mid-90s, less than 3 points above Stanza for tagging, and neck-and-neck for lemmatization, and with human LAS at 84.66, about 8 points above Stanza on average. To illustrate why humans disagree on syntax especially in technical genres, we offer a brief example of parsing a \genre{legal} case law designation for `410 U.S.~113' in Figure \ref{fig:us113}. `410 U.S.' is a volume of US Supreme Court cases, including case `113' (Roe v. Wade) -- one annotator (in black) analyzes `113' (the case) as the head, which is modified by the name of the volume that includes it, while the other treats the volume as the head, with a numerical modifier attached as \texttt{dep}, similar to how GUM annotates cases like `Page 5.' 
Without good intuitions about Supreme Court case nomenclature and very clear guidelines, any chance of perfect agreement is hampered by a myriad of such cases.

On the other hand, some potentially difficult genres, such as \genre{esports}, turned out to have high human agreement for tokenization, tagging and lemmatization, despite well known challenges in annotating User Generated Content (UGC, see \citealt{SanguinettiEtAl2022}).

\paragraph{Entity Recognition and Coreference Resolution} 
For NNER, we evaluate a SOTA neural system (seq2set, \citealt{ijcai2021p0542}). In both full GENTLE and snippets, we consider plain text with gold tokenization as input and use precision, recall and F1 to evaluate. In Table \ref{tab:EVAL4NLP}, F1 drops over 4 points on average, and over 13 points on \genre{legal}. Inspection reveals most errors involve malpredicted spans, especially when deciding entity boundaries with PP attachment, apposition, or coordination. For example, in
\textit{\textcolor{red}{[}Proto-Germanic *nēhwist (“\textcolor{blue}{[}nearest\textcolor{blue}{]$_2$}, \textcolor{orange}{[}closest\textcolor{orange}{]$_3$}”)\textcolor{red}{]$_1$}}, span 2 (\textcolor{blue}{blue}) and span 3 (\textcolor{orange}{orange}) are appositions providing additional information for the word \textit{nēhwist} and span 1 (\textcolor{red}{red}) as a whole forms a non-named entity span, but neither of them are correctly predicted by the model. \genre{proof} outperforms GUM because mathematical variables, which are frequent in \genre{proof}, are easier to identify compared with other types of entities. We also observe this in Table \ref{tab:humanEval}, where IAA is the highest for \genre{proof}. Note that IAA for typed and untyped entities are identical; this is because most entities in \genre{proof}, e.g.~mathematical variables, are \textit{abstract}. 

The coreference resolution task uses MTL-coref \cite{zhu2023incorporating}, a new SOTA model for the GUM benchmark which is trained with singletons and other entity-level information. We use the F1-measure of MUC, B$^3$, CEAF$_{\phi4}$, and the average CoNLL score as evaluation metrics. Table \ref{tab:EVAL4NLP} reveals that the model performs substantially worse on GENTLE, with nearly 32 points degradation. Genre-wise analysis reveals that \genre{dictionary}, which has few pronouns, performs worst, while \genre{threat}, rich in pronouns, scores best in GENTLE. This shows that the model struggles more with complex NPs (with possible PP attachments) and proper nouns but can more easily identify coreference chains involving pronouns (and especially the easy pronouns `I' and `you' in \genre{threat} letters). For instance, in \texttt{GENTLE\_epsorts\_fortnite}, the model incorrectly clusters \textit{\textcolor{red}{[}Kreo\textcolor{red}{]$_1$} ... \textcolor{red}{[}him\textcolor{red}{]$_1$} ... \textcolor{red}{[}Maufin\textcolor{red}{]$_1$}}, a chain including multiple names unseen during training.

\paragraph{RST Segmentation and Parsing} 
We evaluate GENTLE on two RST tasks: elementary discourse unit (EDU) segmentation and RST parsing.
For EDU segmentation, we use DisCoDisCo \citep{gessler-etal-2021-discodisco}, the winning system in the 2021 DISRPT shared task on segmentation.
We evaluate EDU segmentation under two conditions: `Gold', where the full, human-provided UD parses for GENTLE documents are provided to the system; and `Trankit', where with the sole exception of tokenization (which remains human-provided), all UD parse information is provided by Trankit's \citep{nguyen-etal-2021-trankit} default English model.

For RST parsing, we use the best setting from the bottom-up neural parser by \citet{guz-carenini-2020-coreference}, \texttt{SpanBERT-NoCoref}, which obtained the SOTA performance on GUM as of v8 \cite{liu-zeldes-2023-eacl} using the original Parseval procedure on binary trees, following \citet{morey-etal-2017-much}. We evaluate using gold discourse units for simplicity and comparability with previous work. 

Unsurprisingly, GENTLE contains challenging materials even with gold discourse units: overall, the best-performing genre is \genre{legal} while the worst-performing genre is \genre{esports}. By examining dependency conversions of gold vs.~predicted trees following \citet{li-etal-2014-text}, we found that the model was only able to correctly identify the Central Discourse Unit in 6 out of 26 documents (23.1\%) in GENTLE. The top 2 most difficult relation classes are \textsc{Topic} and \textsc{Explanation}, both of which tend to lack explicit and unambiguous cues such as discourse markers, and may require an understanding over multiple EDUs. 

% RESULTS ON VALIDATED DATA 
% CDU Performance: % 23.076923076923077
% GENTLE_threat_malik
% GENTLE_dictionary_next
% GENTLE_dictionary_trust
% GENTLE_poetry_flower
% GENTLE_esports_fifa
% GENTLE_legal_service

% topic 	 0.143
% explanation 	 0.193
% root 	 0.231
% evaluation 	 0.254
% context 	 0.265
% causal 	 0.317
% restatement 	 0.328
% adversative 	 0.395
% mode 	 0.488
% elaboration 	 0.557
% organization 	 0.6
% joint 	 0.636
% contingency 	 0.673
% same-unit 	 0.782
% purpose 	 0.833
% attribution 	 0.843

\section{Conclusion}\label{sec:conclusion}

We have introduced GENTLE, a new, genre-diverse, richly-annotated test corpus for English. 
While this new resource is relatively small, the challenging genres included in the corpus are diverse not only in topic, but also in terms of medium and communicative intent. 
The 8 genres have considerably distinct characteristics reflected in metrics and label distributions for individual annotation layers. 
These genres also differ substantially from the 12 genres in the GUM reference corpus. 
As such, GENTLE serves as an important complement to GUM's test set, and can provide valuable insights into NLP systems' ability to perform on OOD data.

We found in evaluations that system performance generally degraded on GENTLE compared to GUM, corroborating prior findings that NLP systems degrade on OOD data. 
However, degradation was not uniform, and different genres presented differing degrees of difficulty for different NLP tasks. 
For dependency parsing, the steepest degradation was in \genre{syllabus} and \genre{proof},
while entity recognition saw particularly poor performance in \genre{legal} and \genre{dictionary},
and RST parsing performed lowest on \genre{esports}, \genre{dictionary} and \genre{poetry}.
It is thus necessary to have a wide variety of genres available for evaluation if one aims for a holistic understanding of the capabilities and limitations of an NLP system. 

Moreover, it is worth noting that the annotation tasks for our challenge genres were not just difficult for the NLP systems, but for our human annotators as well. 
Our IAA experiments showed that human annotation generally outperformed the NLP systems in terms of accuracy.
However, some genres stood out as being particularly difficult for humans, such as \genre{dictionary}, which suggests that it would be beneficial to develop additional annotation guidelines targeting difficult cases that arise from genre-specific phenomena.

With the introduction of GENTLE and the results from the above evaluation experiments, we hope to encourage the use of genre-diverse test corpora for NLP benchmarks. 
This will allow researchers to obtain realistic measures of how NLP systems will perform on OOD data, which is frequently the use case of interest in real-world applications of NLP technologies.

\section*{Limitations}

Our corpus is designed to serve as a challenge set, and is limited in size: each of the 8 genres ranges from 2k to 2.5k tokens, totaling around 18k tokens. Given the amount of work necessary for multilayer annotations, building a larger challenge set was not deemed realistic with the limited resources available for this project, and is left for future work.

Additionally, the evaluation of inter-annotator agreement is limited to a small amount of data, since double annotating the amount of annotation layers involved is costly. In particular, the evaluation is limited by the use of a common gold tokenization standard to facilitate reporting commonly used scores (Cohen's Kappa, tagging accuracy, NNER F1, etc.), which do not reflect cascading errors due to tokenization disagreements. Additionally, we did not perform double annotation experiments for RST discourse parsing, as these would require annotating entire documents in each genre, which would exceed the amount of data we were able to have annotated for this evaluation.

\bibliography{anthology,custom}
\bibliographystyle{acl_natbib}

\clearpage

\appendix

\section{Genre Descriptions}
\label{sec:genres}

GENTLE comprises 8 genres, with each having 2 to 5 individual documents---cf.~Table \ref{tab:overview-gentle-stats}.
They are as follows:

\begin{itemize}
    \item \textbf{\genre{dictionary}} -- entries for a single English word from Wiktionary (\url{https://en.wiktionary.org}). GENTLE includes documents for the words \textit{next}, \textit{trust}, and \textit{school}.
    \item \textbf{\genre{esports}} -- transcripts of a YouTube video clip containing esport commentary. GENTLE includes two documents: one featuring Fortnite, and the other featuring FIFA 20.
    \item \textbf{\genre{legal}} -- segments of legal text from the United States. Of the two documents, one is a portion of the Supreme Court opinion for Roe v.~Wade (1973) from Wikisource (\url{https://en.wikisource.org}), and the other is a portion of a contract, extracted from the \textbf{C}ontract \textbf{U}nderstanding \textbf{A}tticus \textbf{D}ataset (CUAD) v1 from the The Atticus Project \cite{hendrycks2021cuad}. 
    \item \textbf{\genre{medical}} -- snippets of a Subjective, Objective, Assessment and Plan (SOAP) note. A SOAP note is a common kind of text used by medical professionals to document a patient's medical visits and history. The notes are taken from MTSamples (\url{https://mtsamples.com}). 
    \item \textbf{\genre{poetry}} -- poems taken from Wikisource (\url{https://en.wikisource.org/wiki/Portal:Poetry}). The poems come from 3 different authors and are of varying lengths.
    \item \textbf{\genre{proof}} -- mathematical proofs taken from ProofWiki (\url{https://proofwiki.org}).
    \item \textbf{\genre{syllabus}} -- syllabuses taken from course materials posted publicly on GitHub.
    \item \textbf{\genre{threat}} -- threat letters recorded in publicly available United States court proceedings. Accessed through casetext (\url{https://casetext.com/cases}; see also \citealt{Abrams2019} for some analysis of these texts).
\end{itemize}

\section{Full Label Residual Tables}
\label{appendix:residuals}

The following tables give complete standardized Pearson residuals for label distributions in each GENTLE genre, along with comparisons to GUM as a whole and GUM \genre{news} in particular. Tables \ref{tab:resid-upos}--\ref{tab:resid-rst} give numbers for UPOS, dependency, entity, and RST coarse labels respectively.

\begin{table*}[t!b]
    %\begin{adjustbox}{angle=90}

    \centering
    \small
    \resizebox{\textwidth}{!}{%
    \begin{tabular}{l|rrrrrrrrrrrrrrrrr}
    \toprule
 & \textbf{ADJ} & \textbf{ADP} & \textbf{ADV} & \textbf{AU\textbf{X}} & \textbf{CCONJ} & \textbf{DET} & \textbf{INTJ} & \textbf{NOUN} & \textbf{NUM} & \textbf{PART} & \textbf{PRON} & \textbf{PROPN} & \textbf{PUNCT} & \textbf{SCONJ} & \textbf{SYM} & \textbf{VERB} & \textbf{X} \\
\midrule
GUM &\textcolor{red}{-0.5} &\textcolor{red}{-2.1} &\textcolor{blue}{11.3} &\textcolor{blue}{8.7} &\textcolor{blue}{2.3} &\textcolor{blue}{1.8} &\textcolor{blue}{13.3} &\textcolor{red}{-15.1} &\textcolor{red}{-5.2} &\textcolor{blue}{3.9} &\textcolor{blue}{20.8} &\textcolor{red}{-25.1} &\textcolor{red}{-0.2} &\textcolor{blue}{4.1} &\textcolor{red}{-14.6} &\textcolor{blue}{6.6} &\textcolor{red}{-22.6} \\
GUM$_\textrm{news}$ &\textcolor{red}{-1.3} &\textcolor{blue}{5.9} &\textcolor{red}{-11.5} &\textcolor{red}{-6.2} &\textcolor{red}{-4.8} &\textcolor{blue}{4.9} &\textcolor{red}{-11.5} &\textcolor{blue}{5.7} &\textcolor{blue}{4.6} &\textcolor{red}{-1.1} &\textcolor{red}{-21.7} &\textcolor{blue}{41.4} &\textcolor{red}{-5.4} &\textcolor{red}{-3.3} &\textcolor{red}{-1.9} &\textcolor{red}{-3.3} &\textcolor{red}{-6.6} \\
\midrule
\genre{dictionary} &\textcolor{blue}{7.9} &\textcolor{red}{-1.2} &\textcolor{red}{-5.9} &\textcolor{red}{-7.2} &\textcolor{red}{-0.4} &\textcolor{red}{-5.2} &\textcolor{red}{-2.9} &\textcolor{blue}{4.4} &\textcolor{red}{-5.0} &\textcolor{red}{-0.6} &\textcolor{red}{-8.6} &\textcolor{red}{-4.9} &\textcolor{blue}{20.4} &\textcolor{red}{-4.6} &\textcolor{red}{-2.1} &\textcolor{red}{-6.1} &\textcolor{blue}{21.2} \\
\genre{esports} &\textcolor{red}{-3.3} &\textcolor{red}{-1.1} &\textcolor{blue}{5.9} &\textcolor{blue}{2.1} &\textcolor{red}{-1.9} &\textcolor{red}{-0.6} &\textcolor{blue}{0.4} &\textcolor{red}{-5.2} &\textcolor{blue}{1.8} &\textcolor{blue}{4.7} &\textcolor{blue}{3.7} &\textcolor{blue}{1.4} &\textcolor{red}{-4.1} &\textcolor{red}{-1.5} &\textcolor{red}{-1.3} &\textcolor{blue}{4.2} &\textcolor{red}{-2.1} \\
\genre{legal} &\textcolor{blue}{0.3} &\textcolor{blue}{0.0} &\textcolor{red}{-4.7} &\textcolor{red}{-5.5} &\textcolor{blue}{3.0} &\textcolor{blue}{3.2} &\textcolor{red}{-4.1} &\textcolor{blue}{4.7} &\textcolor{blue}{2.6} &\textcolor{blue}{0.4} &\textcolor{red}{-9.5} &\textcolor{blue}{3.9} &\textcolor{blue}{0.8} &\textcolor{blue}{0.0} &\textcolor{blue}{4.9} &\textcolor{red}{-3.1} &\textcolor{blue}{18.4} \\
\genre{medical} &\textcolor{blue}{7.6} &\textcolor{blue}{0.0} &\textcolor{red}{-5.3} &\textcolor{red}{-0.4} &\textcolor{blue}{0.0} &\textcolor{red}{-5.0} &\textcolor{red}{-4.0} &\textcolor{blue}{11.6} &\textcolor{blue}{5.4} &\textcolor{red}{-4.2} &\textcolor{red}{-2.6} &\textcolor{red}{-6.9} &\textcolor{blue}{1.0} &\textcolor{red}{-3.3} &\textcolor{red}{-0.4} &\textcolor{red}{-4.8} &\textcolor{blue}{8.9} \\
\genre{poetry} &\textcolor{red}{-2.6} &\textcolor{red}{-1.4} &\textcolor{blue}{5.0} &\textcolor{red}{-4.5} &\textcolor{blue}{2.7} &\textcolor{blue}{2.1} &\textcolor{red}{-2.8} &\textcolor{red}{-1.6} &\textcolor{red}{-5.3} &\textcolor{red}{-4.5} &\textcolor{blue}{5.5} &\textcolor{red}{-7.1} &\textcolor{blue}{5.5} &\textcolor{blue}{0.9} &\textcolor{red}{-1.9} &\textcolor{blue}{1.9} &\textcolor{red}{-2.2} \\
\genre{proof} &\textcolor{red}{-1.0} &\textcolor{blue}{0.4} &\textcolor{blue}{0.5} &\textcolor{blue}{0.3} &\textcolor{red}{-1.2} &\textcolor{red}{-4.7} &\textcolor{red}{-3.9} &\textcolor{blue}{14.5} &\textcolor{blue}{1.8} &\textcolor{red}{-5.8} &\textcolor{red}{-9.0} &\textcolor{red}{-8.1} &\textcolor{blue}{0.9} &\textcolor{blue}{3.7} &\textcolor{blue}{56.3} &\textcolor{red}{-6.4} &\textcolor{red}{-2.3} \\
\genre{syllabus} &\textcolor{red}{-1.4} &\textcolor{red}{-3.1} &\textcolor{red}{-6.6} &\textcolor{red}{-5.1} &\textcolor{blue}{2.5} &\textcolor{red}{-6.4} &\textcolor{red}{-2.5} &\textcolor{blue}{13.6} &\textcolor{blue}{7.3} &\textcolor{red}{-4.5} &\textcolor{red}{-10.4} &\textcolor{blue}{11.2} &\textcolor{red}{-3.4} &\textcolor{red}{-3.3} &\textcolor{blue}{5.4} &\textcolor{red}{-5.3} &\textcolor{blue}{54.5} \\
\genre{threat} &\textcolor{red}{-2.5} &\textcolor{red}{-1.3} &\textcolor{blue}{0.8} &\textcolor{blue}{5.5} &\textcolor{red}{-0.5} &\textcolor{red}{-2.2} &\textcolor{blue}{1.6} &\textcolor{red}{-2.6} &\textcolor{red}{-2.1} &\textcolor{blue}{3.3} &\textcolor{blue}{13.2} &\textcolor{red}{-7.4} &\textcolor{red}{-7.0} &\textcolor{blue}{2.5} &\textcolor{red}{-1.7} &\textcolor{blue}{4.5} &\textcolor{red}{-0.9}
\\ 
\bottomrule
    \end{tabular}
    }
\caption{Residuals for UPOS Labels by Genre. }\label{tab:resid-upos}
\end{table*}

% % SUBTABLE 

\begin{table*}[t!b]
\centering
\small
\subfloat[Part 1.]{%
\resizebox{\textwidth}{!}{%
\begin{tabular}{l|rrrrrrrrrrrrrrrrrr}
\toprule
 & \textbf{acl} & \textbf{advcl} & \textbf{advmod} & \textbf{amod} & \textbf{appos} & \textbf{aux} & \textbf{case} & \textbf{cc} & \textbf{cc}omp & \textbf{compound} & \textbf{conj} & \textbf{cop} & \textbf{csubj} & \textbf{dep} & \textbf{det} & \textbf{discourse} & \textbf{dislocated} & \textbf{expl} \\
\midrule
GUM &\textcolor{blue}{1.3} &\textcolor{blue}{4.3} &\textcolor{blue}{12.2} &\textcolor{red}{-1.9} &\textcolor{red}{-12.8} &\textcolor{blue}{4.7} &\textcolor{red}{-4.3} &\textcolor{blue}{2.3} &\textcolor{red}{-1.2} &\textcolor{red}{-16.1} &\textcolor{red}{-2.2} &\textcolor{blue}{7.8} &\textcolor{blue}{3.8} &\textcolor{red}{-20.9} &\textcolor{blue}{1.4} &\textcolor{blue}{9.9} &\textcolor{red}{-2.9} &\textcolor{blue}{4.3} \\
GUM$_\textrm{news}$ &\textcolor{red}{-0.6} &\textcolor{red}{-2.6} &\textcolor{red}{-11.7} &\textcolor{blue}{3.5} &\textcolor{blue}{7.5} &\textcolor{red}{-2.1} &\textcolor{blue}{9.6} &\textcolor{red}{-5.1} &\textcolor{blue}{4.9} &\textcolor{blue}{20.3} &\textcolor{red}{-4.5} &\textcolor{red}{-7.1} &\textcolor{red}{-3.1} &\textcolor{red}{-2.5} &\textcolor{blue}{5.3} &\textcolor{red}{-9.3} &\textcolor{red}{-1.4} &\textcolor{red}{-4.0} \\
\midrule
\genre{dictionary} &\textcolor{red}{-1.6} &\textcolor{red}{-4.4} &\textcolor{red}{-6.3} &\textcolor{blue}{2.1} &\textcolor{blue}{7.0} &\textcolor{red}{-5.2} &\textcolor{red}{-2.4} &\textcolor{red}{-0.5} &\textcolor{red}{-4.1} &\textcolor{red}{-2.2} &\textcolor{blue}{9.7} &\textcolor{red}{-4.7} &\textcolor{red}{-1.7} &\textcolor{blue}{5.6} &\textcolor{red}{-5.1} &\textcolor{red}{-3.1} &\textcolor{red}{-0.3} &\textcolor{red}{-2.4} \\
\genre{esports} &\textcolor{red}{-1.5} &\textcolor{blue}{0.9} &\textcolor{blue}{4.7} &\textcolor{red}{-3.9} &\textcolor{red}{-2.3} &\textcolor{blue}{2.3} &\textcolor{red}{-3.0} &\textcolor{red}{-2.4} &\textcolor{red}{-0.5} &\textcolor{red}{-0.4} &\textcolor{red}{-0.3} &\textcolor{blue}{0.5} &\textcolor{red}{-1.3} &\textcolor{red}{-2.2} &\textcolor{red}{-0.7} &\textcolor{blue}{2.1} &\textcolor{blue}{9.2} &\textcolor{red}{-1.2} \\
\genre{legal} &\textcolor{blue}{4.2} &\textcolor{red}{-1.3} &\textcolor{red}{-4.4} &\textcolor{blue}{1.4} &\textcolor{blue}{3.6} &\textcolor{red}{-4.1} &\textcolor{blue}{1.8} &\textcolor{blue}{2.7} &\textcolor{red}{-2.7} &\textcolor{blue}{3.4} &\textcolor{blue}{1.8} &\textcolor{red}{-3.4} &\textcolor{red}{-1.4} &\textcolor{blue}{17.2} &\textcolor{blue}{3.3} &\textcolor{red}{-3.4} &\textcolor{red}{-0.2} &\textcolor{red}{-1.7} \\
\genre{medical} &\textcolor{red}{-3.8} &\textcolor{red}{-4.3} &\textcolor{red}{-5.9} &\textcolor{blue}{7.7} &\textcolor{red}{-1.3} &\textcolor{blue}{0.4} &\textcolor{blue}{0.2} &\textcolor{blue}{0.1} &\textcolor{red}{-1.8} &\textcolor{blue}{1.9} &\textcolor{blue}{3.5} &\textcolor{red}{-0.9} &\textcolor{red}{-1.6} &\textcolor{blue}{7.7} &\textcolor{red}{-4.7} &\textcolor{red}{-3.3} &\textcolor{red}{-0.2} &\textcolor{red}{-1.2} \\
\genre{poetry} &\textcolor{blue}{1.9} &\textcolor{blue}{1.5} &\textcolor{blue}{3.8} &\textcolor{red}{-2.8} &\textcolor{red}{-2.0} &\textcolor{red}{-4.3} &\textcolor{red}{-1.6} &\textcolor{blue}{2.5} &\textcolor{blue}{1.3} &\textcolor{red}{-5.7} &\textcolor{red}{-0.3} &\textcolor{red}{-1.8} &\textcolor{red}{-0.7} &\textcolor{red}{-3.6} &\textcolor{blue}{2.2} &\textcolor{red}{-1.9} &\textcolor{blue}{1.7} &\textcolor{red}{-0.7} \\
\genre{proof} &\textcolor{red}{-1.2} &\textcolor{red}{-0.3} &\textcolor{red}{-0.1} &\textcolor{red}{-3.0} &\textcolor{blue}{3.6} &\textcolor{red}{-3.2} &\textcolor{blue}{1.6} &\textcolor{blue}{0.5} &\textcolor{blue}{1.6} &\textcolor{red}{-6.1} &\textcolor{blue}{3.1} &\textcolor{blue}{4.6} &\textcolor{blue}{2.6} &\textcolor{blue}{10.9} &\textcolor{red}{-5.2} &\textcolor{red}{-3.3} &\textcolor{blue}{3.0} &\textcolor{blue}{3.9} \\
\genre{syllabus} &\textcolor{red}{-3.6} &\textcolor{red}{-2.9} &\textcolor{red}{-7.7} &\textcolor{red}{-0.5} &\textcolor{blue}{19.8} &\textcolor{red}{-2.8} &\textcolor{red}{-3.5} &\textcolor{blue}{2.7} &\textcolor{red}{-3.8} &\textcolor{blue}{16.6} &\textcolor{blue}{1.7} &\textcolor{red}{-4.5} &\textcolor{red}{-1.7} &\textcolor{blue}{46.3} &\textcolor{red}{-6.5} &\textcolor{red}{-2.0} &\textcolor{red}{-0.3} &\textcolor{red}{-2.4} \\
\genre{threat} &\textcolor{blue}{2.5} &\textcolor{blue}{2.4} &\textcolor{blue}{3.0} &\textcolor{red}{-3.6} &\textcolor{red}{-2.8} &\textcolor{blue}{5.9} &\textcolor{red}{-2.2} &\textcolor{red}{-0.7} &\textcolor{blue}{2.2} &\textcolor{red}{-2.8} &\textcolor{red}{-0.0} &\textcolor{blue}{1.0} &\textcolor{blue}{0.4} &\textcolor{red}{-1.9} &\textcolor{red}{-2.1} &\textcolor{blue}{3.3} &\textcolor{blue}{1.7} &\textcolor{blue}{0.5} \\
\bottomrule
\end{tabular}
}}
\vspace{11pt}

\quad\quad

\subfloat[Part 2.]{%
\resizebox{\textwidth}{!}{%
\begin{tabular}{l|rrrrrrrrrrrrrrrrrr}
\toprule
& \textbf{fixed} & \textbf{flat} & \textbf{goeswith} & i\textbf{obj} & \textbf{list} & \textbf{mark} & \textbf{nmod} & \textbf{nsubj} & \textbf{nummod} & \textbf{obj} & \textbf{obl} & \textbf{orphan} & \textbf{parataxis} & \textbf{punct} & \textbf{reparandum} & \textbf{root} & \textbf{vocative} & \textbf{xcomp} \\
\midrule
GUM
&\textcolor{blue}{0.0} &\textcolor{red}{-9.7} &\textcolor{red}{-2.7} &\textcolor{blue}{0.6} &\textcolor{blue}{1.8} &\textcolor{blue}{5.0} &\textcolor{red}{-4.0} &\textcolor{blue}{8.4} &\textcolor{red}{-7.2} &\textcolor{blue}{4.6} &\textcolor{red}{-2.2} &\textcolor{red}{-0.1} &\textcolor{red}{-4.1} &\textcolor{red}{-0.1} &\textcolor{blue}{5.5} &\textcolor{red}{-2.7} &\textcolor{blue}{1.8} &\textcolor{blue}{4.9} \\
GUM$_\textrm{news}$ &\textcolor{blue}{1.1} &\textcolor{blue}{21.7} &\textcolor{red}{-0.5} &\textcolor{red}{-0.6} &\textcolor{red}{-1.6} &\textcolor{red}{-4.2} &\textcolor{blue}{7.3} &\textcolor{red}{-4.2} &\textcolor{blue}{4.2} &\textcolor{red}{-4.0} &\textcolor{blue}{3.6} &\textcolor{red}{-0.8} &\textcolor{red}{-6.5} &\textcolor{red}{-5.4} &\textcolor{red}{-5.7} &\textcolor{red}{-6.8} &\textcolor{red}{-2.4} &\textcolor{red}{-4.8} \\
\midrule
\genre{dictionary} 
&\textcolor{red}{-0.3} &\textcolor{red}{-4.5} &\textcolor{blue}{0.5} &\textcolor{red}{-1.7} &\textcolor{red}{-0.4} &\textcolor{red}{-2.2} &\textcolor{red}{-3.1} &\textcolor{red}{-9.8} &\textcolor{red}{-2.9} &\textcolor{red}{-5.3} &\textcolor{red}{-1.7} &\textcolor{red}{-0.0} &\textcolor{blue}{18.4} &\textcolor{blue}{20.4} &\textcolor{red}{-2.0} &\textcolor{blue}{7.3} &\textcolor{red}{-0.3} &\textcolor{red}{-3.9} \\
\genre{esports} &\textcolor{blue}{1.5} &\textcolor{red}{-2.4} &\textcolor{blue}{1.7} &\textcolor{blue}{1.5} &\textcolor{red}{-0.3} &\textcolor{blue}{3.9} &\textcolor{red}{-5.6} &\textcolor{blue}{2.4} &\textcolor{blue}{3.5} &\textcolor{blue}{1.0} &\textcolor{blue}{1.8} &\textcolor{red}{-0.4} &\textcolor{blue}{16.7} &\textcolor{red}{-4.1} &\textcolor{blue}{5.8} &\textcolor{red}{-1.5} &\textcolor{red}{-0.2} &\textcolor{blue}{5.2} \\
\genre{legal}
&\textcolor{red}{-0.2} &\textcolor{red}{-1.6} &\textcolor{blue}{0.6} &\textcolor{red}{-1.1} &\textcolor{red}{-0.3} &\textcolor{red}{-0.9} &\textcolor{blue}{3.5} &\textcolor{red}{-6.6} &\textcolor{blue}{0.6} &\textcolor{red}{-0.3} &\textcolor{red}{-1.5} &\textcolor{red}{-0.5} &\textcolor{red}{-2.0} &\textcolor{blue}{0.8} &\textcolor{red}{-1.9} &\textcolor{red}{-2.4} &\textcolor{red}{-0.7} &\textcolor{red}{-2.6} \\
\genre{medical}
&\textcolor{red}{-1.2} &\textcolor{red}{-3.8} &\textcolor{blue}{0.6} &\textcolor{red}{-1.5} &\textcolor{red}{-0.3} &\textcolor{red}{-3.8} &\textcolor{blue}{3.4} &\textcolor{red}{-2.1} &\textcolor{blue}{7.9} &\textcolor{red}{-2.6} &\textcolor{blue}{0.5} &\textcolor{blue}{2.2} &\textcolor{red}{-1.3} &\textcolor{blue}{0.9} &\textcolor{red}{-1.8} &\textcolor{blue}{6.7} &\textcolor{red}{-0.7} &\textcolor{red}{-1.8} \\
\genre{poetry}
&\textcolor{red}{-0.7} &\textcolor{red}{-2.8} &\textcolor{blue}{0.6} &\textcolor{blue}{1.6} &\textcolor{red}{-0.3} &\textcolor{red}{-2.2} &\textcolor{red}{-0.6} &\textcolor{blue}{0.8} &\textcolor{red}{-2.2} &\textcolor{blue}{0.1} &\textcolor{blue}{0.5} &\textcolor{red}{-0.4} &\textcolor{blue}{1.4} &\textcolor{blue}{5.5} &\textcolor{red}{-0.1} &\textcolor{blue}{0.1} &\textcolor{blue}{1.7} &\textcolor{red}{-0.8} \\
\genre{proof}
&\textcolor{red}{-0.3} &\textcolor{red}{-4.5} &\textcolor{blue}{0.6} &\textcolor{red}{-1.5} &\textcolor{red}{-0.3} &\textcolor{blue}{0.0} &\textcolor{blue}{2.7} &\textcolor{blue}{1.4} &\textcolor{red}{-1.2} &\textcolor{red}{-1.2} &\textcolor{blue}{2.1} &\textcolor{blue}{0.1} &\textcolor{red}{-2.3} &\textcolor{blue}{0.9} &\textcolor{red}{-1.8} &\textcolor{blue}{1.3} &\textcolor{red}{-0.6} &\textcolor{red}{-1.1} \\
\genre{syllabus}
&\textcolor{red}{-0.3} &\textcolor{blue}{1.1} &\textcolor{blue}{0.5} &\textcolor{red}{-1.7} &\textcolor{red}{-0.4} &\textcolor{red}{-4.3} &\textcolor{red}{-3.1} &\textcolor{red}{-9.1} &\textcolor{blue}{8.5} &\textcolor{red}{-1.9} &\textcolor{red}{-2.9} &\textcolor{blue}{1.9} &\textcolor{red}{-0.5} &\textcolor{red}{-3.5} &\textcolor{red}{-1.7} &\textcolor{blue}{16.4} &\textcolor{red}{-0.7} &\textcolor{red}{-3.9} \\
\genre{threat}
&\textcolor{red}{-1.5} &\textcolor{red}{-3.3} &\textcolor{blue}{6.0} &\textcolor{blue}{4.0} &\textcolor{red}{-0.3} &\textcolor{blue}{2.6} &\textcolor{red}{-1.3} &\textcolor{blue}{4.6} &\textcolor{blue}{0.9} &\textcolor{blue}{4.2} &\textcolor{blue}{0.4} &\textcolor{red}{-0.4} &\textcolor{blue}{1.1} &\textcolor{red}{-7.0} &\textcolor{red}{-1.6} &\textcolor{red}{-1.1} &\textcolor{blue}{1.6} &\textcolor{blue}{4.1} \\
\bottomrule
\end{tabular}
}}
%\caption{second}
\caption{Residuals for Deprel Labels by Genre.}
\label{tab:resid-dep}
\end{table*}

\begin{table*}[t!b]

    \centering
    \small
    \begin{tabular}{l|rrrrrrrrrr}
    \toprule
 & \textbf{abstract} & \textbf{animal} & \textbf{event} & \textbf{object} & \textbf{organization} & \textbf{person} & \textbf{place} & \textbf{plant} & \textbf{substance} & \textbf{time} \\
\midrule
GUM &\textcolor{red}{-13.8} &\textcolor{blue}{3.5} &\textcolor{red}{-3.1} &\textcolor{blue}{6.4} &\textcolor{red}{-19.5} &\textcolor{blue}{13.4} &\textcolor{blue}{8.7} &\textcolor{blue}{6.4} &\textcolor{blue}{4.2} &\textcolor{red}{-2.3} \\
GUM$_\textrm{news}$ &\textcolor{red}{-15.2} &\textcolor{red}{-7.7} &\textcolor{blue}{7.2} &\textcolor{blue}{0.5} &\textcolor{blue}{30.6} &\textcolor{red}{-6.7} &\textcolor{blue}{3.9} &\textcolor{red}{-2.2} &\textcolor{red}{-0.3} &\textcolor{blue}{5.7} \\
\midrule
\genre{dictionary} &\textcolor{blue}{17.1} &\textcolor{blue}{5.0} &\textcolor{red}{-4.9} &\textcolor{red}{-5.9} &\textcolor{blue}{6.6} &\textcolor{red}{-10.1} &\textcolor{red}{-6.1} &\textcolor{red}{-2.4} &\textcolor{red}{-3.6} &\textcolor{blue}{0.4} \\
\genre{esports} &\textcolor{red}{-9.3} &\textcolor{red}{-2.6} &\textcolor{blue}{9.9} &\textcolor{blue}{0.1} &\textcolor{red}{-0.5} &\textcolor{blue}{7.8} &\textcolor{red}{-2.4} &\textcolor{red}{-2.0} &\textcolor{red}{-3.0} &\textcolor{red}{-0.7} \\
\genre{legal} &\textcolor{blue}{9.8} &\textcolor{red}{-2.7} &\textcolor{blue}{1.9} &\textcolor{red}{-6.7} &\textcolor{blue}{12.6} &\textcolor{red}{-8.0} &\textcolor{red}{-4.5} &\textcolor{red}{-2.1} &\textcolor{red}{-3.9} &\textcolor{blue}{0.2} \\
\genre{medical} &\textcolor{blue}{4.4} &\textcolor{red}{-0.4} &\textcolor{blue}{2.8} &\textcolor{blue}{6.2} &\textcolor{red}{-6.1} &\textcolor{red}{-4.8} &\textcolor{red}{-7.8} &\textcolor{red}{-2.3} &\textcolor{blue}{11.0} &\textcolor{red}{-0.2} \\
\genre{poetry} &\textcolor{red}{-3.6} &\textcolor{blue}{17.8} &\textcolor{red}{-4.3} &\textcolor{blue}{1.5} &\textcolor{red}{-5.4} &\textcolor{blue}{4.2} &\textcolor{blue}{1.3} &\textcolor{red}{-0.8} &\textcolor{red}{-1.9} &\textcolor{red}{-1.4} \\
\genre{proof} &\textcolor{blue}{39.2} &\textcolor{red}{-3.0} &\textcolor{red}{-7.0} &\textcolor{red}{-7.7} &\textcolor{red}{-6.4} &\textcolor{red}{-16.5} &\textcolor{red}{-9.2} &\textcolor{red}{-2.4} &\textcolor{red}{-4.3} &\textcolor{red}{-6.7} \\
\genre{syllabus} &\textcolor{blue}{25.7} &\textcolor{red}{-3.4} &\textcolor{red}{-2.5} &\textcolor{red}{-8.6} &\textcolor{red}{-4.7} &\textcolor{red}{-12.1} &\textcolor{red}{-7.3} &\textcolor{red}{-2.7} &\textcolor{red}{-4.7} &\textcolor{blue}{4.6} \\
\genre{threat} &\textcolor{red}{-4.1} &\textcolor{red}{-2.2} &\textcolor{red}{-1.6} &\textcolor{red}{-1.0} &\textcolor{red}{-2.9} &\textcolor{blue}{12.9} &\textcolor{red}{-2.9} &\textcolor{red}{-2.1} &\textcolor{red}{-3.7} &\textcolor{red}{-3.2} \\
\bottomrule
\end{tabular}
\caption{Residuals for Entity Labels by Genre.}\label{tab:resid-ent}
\end{table*}

\begin{table*}[t!b]

    \centering
    \small
    \resizebox{\textwidth}{!}{%
    \begin{tabular}{l|rrrrrrrrrrrrrrr}
    \toprule
 & \textbf{adversative} & \textbf{attribution} & \textbf{causal} & \textbf{context} & \textbf{contingency} & \textbf{elaboration} & \textbf{evaluation} & \textbf{explanation} & \textbf{joint} & \textbf{mode} & \textbf{organization} & \textbf{purpose} & \textbf{restatement} & \textbf{same} & \textbf{topic} \\
\midrule
GUM &\textcolor{blue}{5.9} &\textcolor{red}{-3.7} &\textcolor{blue}{2.6} &\textcolor{blue}{0.6} &\textcolor{blue}{1.3} &\textcolor{red}{-2.2} &\textcolor{blue}{5.7} &\textcolor{blue}{2.6} &\textcolor{red}{-8.9} &\textcolor{blue}{1.9} &\textcolor{red}{-2.4} &\textcolor{blue}{1.8} &\textcolor{blue}{4.1} &\textcolor{red}{-0.8} &\textcolor{blue}{7.7} \\
GUM$_\textrm{news}$ &\textcolor{red}{-3.6} &\textcolor{blue}{12.1} &\textcolor{blue}{2.1} &\textcolor{blue}{3.6} &\textcolor{red}{-1.8} &\textcolor{blue}{5.3} &\textcolor{red}{-4.4} &\textcolor{red}{-6.0} &\textcolor{red}{-2.7} &\textcolor{red}{-2.1} &\textcolor{red}{-4.6} &\textcolor{blue}{2.2} &\textcolor{red}{-3.5} &\textcolor{blue}{1.4} &\textcolor{red}{-5.2} \\
\midrule
\genre{dictionary} &\textcolor{red}{-3.6} &\textcolor{red}{-5.4} &\textcolor{red}{-4.3} &\textcolor{blue}{1.0} &\textcolor{red}{-2.3} &\textcolor{blue}{2.6} &\textcolor{red}{-4.0} &\textcolor{blue}{0.6} &\textcolor{blue}{3.2} &\textcolor{red}{-1.8} &\textcolor{blue}{5.7} &\textcolor{red}{-2.5} &\textcolor{blue}{0.1} &\textcolor{blue}{3.8} &\textcolor{red}{-2.6} \\
\genre{esports} &\textcolor{blue}{0.3} &\textcolor{red}{-1.2} &\textcolor{red}{-0.2} &\textcolor{blue}{1.5} &\textcolor{red}{-1.6} &\textcolor{red}{-3.2} &\textcolor{blue}{6.8} &\textcolor{red}{-1.2} &\textcolor{blue}{1.3} &\textcolor{red}{-0.2} &\textcolor{red}{-0.8} &\textcolor{blue}{0.9} &\textcolor{blue}{0.3} &\textcolor{red}{-0.3} &\textcolor{red}{-1.4} \\
\genre{legal} &\textcolor{red}{-2.5} &\textcolor{red}{-3.6} &\textcolor{red}{-2.3} &\textcolor{red}{-3.7} &\textcolor{blue}{0.3} &\textcolor{blue}{3.1} &\textcolor{red}{-3.0} &\textcolor{blue}{0.9} &\textcolor{blue}{2.3} &\textcolor{red}{-1.4} &\textcolor{blue}{2.1} &\textcolor{blue}{1.5} &\textcolor{blue}{0.2} &\textcolor{blue}{2.9} &\textcolor{red}{-1.9} \\
\genre{medical} &\textcolor{red}{-1.6} &\textcolor{red}{-2.2} &\textcolor{red}{-2.4} &\textcolor{red}{-3.6} &\textcolor{red}{-1.5} &\textcolor{red}{-4.2} &\textcolor{red}{-3.2} &\textcolor{red}{-3.5} &\textcolor{blue}{12.9} &\textcolor{red}{-1.4} &\textcolor{blue}{9.1} &\textcolor{red}{-2.5} &\textcolor{red}{-1.9} &\textcolor{red}{-3.3} &\textcolor{red}{-2.0} \\
\genre{poetry} &\textcolor{blue}{2.4} &\textcolor{blue}{1.7} &\textcolor{blue}{2.7} &\textcolor{blue}{0.3} &\textcolor{red}{-2.2} &\textcolor{red}{-0.3} &\textcolor{red}{-0.1} &\textcolor{red}{-1.9} &\textcolor{red}{-3.1} &\textcolor{blue}{5.5} &\textcolor{red}{-3.0} &\textcolor{red}{-2.3} &\textcolor{blue}{2.7} &\textcolor{blue}{2.5} &\textcolor{blue}{0.0} \\
\genre{proof} &\textcolor{red}{-4.2} &\textcolor{blue}{1.6} &\textcolor{red}{-2.1} &\textcolor{blue}{2.4} &\textcolor{blue}{4.2} &\textcolor{red}{-1.4} &\textcolor{red}{-3.0} &\textcolor{blue}{8.5} &\textcolor{red}{-0.0} &\textcolor{blue}{0.4} &\textcolor{blue}{2.7} &\textcolor{red}{-2.5} &\textcolor{red}{-2.3} &\textcolor{red}{-3.0} &\textcolor{red}{-2.0} \\
\genre{syllabus} &\textcolor{red}{-3.3} &\textcolor{red}{-5.1} &\textcolor{red}{-4.3} &\textcolor{red}{-5.5} &\textcolor{red}{-0.5} &\textcolor{red}{-0.7} &\textcolor{red}{-4.1} &\textcolor{red}{-4.0} &\textcolor{blue}{18.2} &\textcolor{red}{-1.8} &\textcolor{blue}{4.3} &\textcolor{red}{-2.7} &\textcolor{red}{-3.5} &\textcolor{red}{-2.0} &\textcolor{red}{-2.6} \\
\genre{threat} &\textcolor{blue}{1.2} &\textcolor{blue}{1.1} &\textcolor{blue}{0.5} &\textcolor{red}{-2.2} &\textcolor{blue}{4.1} &\textcolor{red}{-0.8} &\textcolor{blue}{2.4} &\textcolor{blue}{6.4} &\textcolor{red}{-1.3} &\textcolor{red}{-0.5} &\textcolor{red}{-2.9} &\textcolor{red}{-0.3} &\textcolor{red}{-1.3} &\textcolor{red}{-1.5} &\textcolor{red}{-2.0} \\
\bottomrule  
\end{tabular}
}
\caption{Residuals for RST Relation Classes by Genre.}\label{tab:resid-rst}
\end{table*}

\end{document}